\documentclass{article}

\PassOptionsToPackage{numbers, compress}{natbib}

\usepackage[preprint]{neurips_2026}

\usepackage[utf8]{inputenc} 
\usepackage[T1]{fontenc}    
\usepackage{hyperref}       
\usepackage{url}            
\usepackage{booktabs}       
\usepackage{colortbl}       
\usepackage{amsfonts}       
\usepackage{nicefrac}       
\usepackage{microtype}      
\usepackage{xcolor}         

\usepackage{multirow}
\usepackage{graphicx}
\usepackage{amsmath}
\usepackage[ruled,vlined,linesnumbered]{algorithm2e}
\SetArgSty{textnormal}
\usepackage[capitalize,noabbrev]{cleveref}
\usepackage{subcaption}

\usepackage{enumitem}

\usepackage{listings}
\definecolor{codegreen}{rgb}{0,0.6,0}
\definecolor{codegray}{rgb}{0.5,0.5,0.5}
\definecolor{codepurple}{rgb}{0.58,0,0.82}
\definecolor{backcolour}{rgb}{0.95,0.95,0.92}
\newcommand{\audiodescbench}{StrAD}
\newcommand{\audiodesczeromodel}{StrAD-Zero}
\newcommand{\audiodescmodel}{StrAD-FT}
\newcommand{\up}{\ensuremath{\uparrow}}
\newcommand{\down}{\ensuremath{\downarrow}}

\title{StrAD: A Streaming Method and Benchmark for Audio Description Generation for Long-form Videos}

\author{%
 Julian Spravil\textsuperscript{\rm 1,2,}\thanks{Corresponding author: \texttt{julian.spravil\,@\,iais.fraunhofer.de}} \quad
 Sebastian Houben\textsuperscript{\rm 1,3} \quad
 Sven Behnke\textsuperscript{\rm 1,2,4,5} \\[1ex]
 \textsuperscript{1}Fraunhofer IAIS, Germany \\
 \textsuperscript{2}AIS, Computer Science Institute VI, University of Bonn, Germany \\
 \textsuperscript{3}A2S, University of Applied Sciences Bonn-Rhein-Sieg, Germany \\
 \textsuperscript{4}Lamarr Institute for Machine Learning and Artificial Intelligence, Germany \\
 \textsuperscript{5}Center for Robotics, University of Bonn, Germany \\
}

\begin{document}

\maketitle

\begin{abstract}
Visual content is the dominant medium of communication, yet without audio descriptions (ADs), it remains inaccessible to blind and low-vision people.
ADs narrate context-relevant visual events during natural audio pauses.
Manually creating ADs is expensive, limiting coverage to a small fraction of available content.
Most existing automatic AD generation methods frame the task as video clip captioning, requiring ground-truth timestamps and additional context cues such as character databases.
Current benchmarks reinforce this framing, consisting of short video segments paired with automatic or task-mismatched annotations.
We introduce \audiodescbench{}, a benchmark for long-form AD generation on full-length videos spanning diverse genres such as movies, documentaries, short films, performances, and video games.
We reformulate AD generation as streaming dense video captioning.
Our approach processes full-length videos with a sliding window, inserting ADs into existing transcripts without ground-truth timestamps, and supports both fine-tuned models and zero-shot prompting of vision-language models.
On the segment-level task with given timestamps, our fine-tuned \audiodescmodel{} sets the state of the art on CMD-AD with 36.3\,CIDEr (+10.0 over Shot-by-shot), establishes a reference point on \audiodescbench{} (51.0\,CIDEr), and remains competitive on MAD-Eval at 24.9\,CIDEr.
On the full-video streaming task, \audiodescmodel{} reaches a SODA score of 2.4 against 1.1 for our zero-shot baseline \audiodesczeromodel{}, though both exhibit limitations in temporal localization and narrative coherence.
While prior work has tackled full-video AD generation in an offline, multi-stage fashion, ours is the first streaming approach, generating ADs on the fly without ground-truth timestamps.
\audiodescbench{} makes progress on full-video AD generation measurable, a prerequisite for scaling accessibility.
\end{abstract}

\section{Introduction}\label{sec:introduction}
Audio descriptions (ADs) are short narrations placed in speech gaps to convey on-screen events without overlapping dialogue or other relevant sounds~\citep{remael2015pictures}.
For the estimated 2.2 billion people worldwide with visual impairments~\citep{who2019vision}, ADs are an important visual aid and sometimes the only way to follow visual media.
Today, trained describers write ADs manually.
The process is slow and expensive, leaving the vast majority of video content, such as online videos, without ADs.
Interactive video games are especially demanded yet hard to describe~\citep{larreina2024games}.

Automatic AD generation has therefore received growing interest.
An early approach by \citet{lakritz2006semi} leverages screenplays to semi-automatically generate AD\@.
Later, \citet{wang2021toward} proposed an automatic, full-video AD generation solution for the first time.
More recent methods focus mostly on single-sentence AD generation by fine-tuning vision-language models (VLMs)~\citep{han2023autoad, han2023autoad2, han2024autoad3, lin2024learning, wang2025uniad, fang2025distinctad} or applying large language models (LLMs) in a zero-shot manner~\citep{xie2024autoadzero, chu2024llm, xie2025shotbyshot, park2025narrad, khandelwal2025coherentad, zhang2024mmnarrator}.
These approaches share three key limitations:
(i) Framing the problem as a single-sentence AD generation task for a video excerpt, as defined by ground-truth start and end timestamps~\citep{han2023autoad, han2023autoad2, han2024autoad3, xie2024autoadzero, xie2025shotbyshot,wang2025uniad,park2025narrad}.
This makes the task similar to video clip captioning, but with extra context, such as a character database.
In a real-world setting, timestamps and extra context are usually unavailable.
A recent user study also highlights the potential challenges of scaling clip-level AD to full videos~\citep{braun2025evaluation}.
(ii) Most methods ignore the audio signal entirely.
Vision-only generation overlooks the role of audio in supporting AD generation without being overly verbose: tone of voice, ambient sounds, and dialogue can already resolve story ambiguities.
(iii) Complex AD forms are not considered, such as very short, fragmented, or multi-sentence descriptions.
One reason for these limitations is the lack of benchmarks to evaluate the full video setting.
Current evaluation datasets are restricted to short clips from movies and TV series, with annotations either verified but task-misaligned (MAD-Eval~\citep{soldan2022MAD,han2023autoad}) or unchecked (CMD-AD~\citep{han2024autoad3}, TV-AD~\citep{xie2024autoadzero}).
To quantify performance in the full video setting, extensive user studies are currently required~\citep{wang2021toward}.
While user studies are important for ensuring that user requirements are taken into account~\citep{braun2025evaluation}, the lack of open access weights and source code of methods, mostly due to copyright reasons, makes progress tracking impossible.

\begin{figure}[!t]
    \centering
    \begin{subfigure}[t]{0.32\textwidth}
        \centering
        \includegraphics[width=\linewidth]{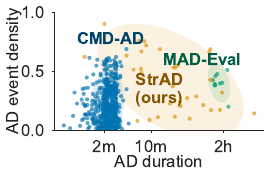}
        \caption{Duration \& Density.}
        \label{fig:event-cover}
    \end{subfigure}%
    \hfill
    \begin{subfigure}[t]{0.32\textwidth}
        \centering
        \includegraphics[width=\linewidth]{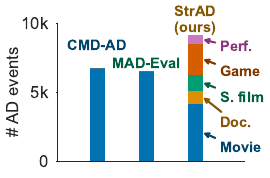}
        \caption{Genre Diversity.}
        \label{fig:genres}
    \end{subfigure}%
    \hfill
    \begin{subfigure}[t]{0.32\textwidth}
        \centering
        \includegraphics[width=\linewidth]{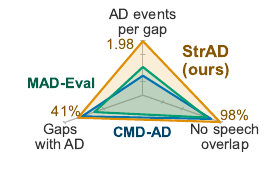}
        \caption{AD Quality.}
        \label{fig:ad-duration}
    \end{subfigure}
    \caption{\textbf{Statistics of \audiodescbench{}.} Relative to the MAD-Eval~\citep{soldan2022MAD, han2023autoad} and CMD-AD test split~\citep{bain2020CMD, han2024autoad3}, \audiodescbench{} (\subref{fig:event-cover})~spans a wider range of video durations and AD densities; (\subref{fig:genres})~covers diverse formats including documentaries (Doc.), short films (S. Film), video games, and performances (Perf.); and (\subref{fig:ad-duration})~provides more speech gaps with AD, more AD events per gap, and less speech overlap.}
    \label{fig:stats}
\end{figure}

\begin{figure}[t]
    \centering
    \begin{subfigure}[t]{0.32\textwidth}
        \centering
        \includegraphics[width=\linewidth]{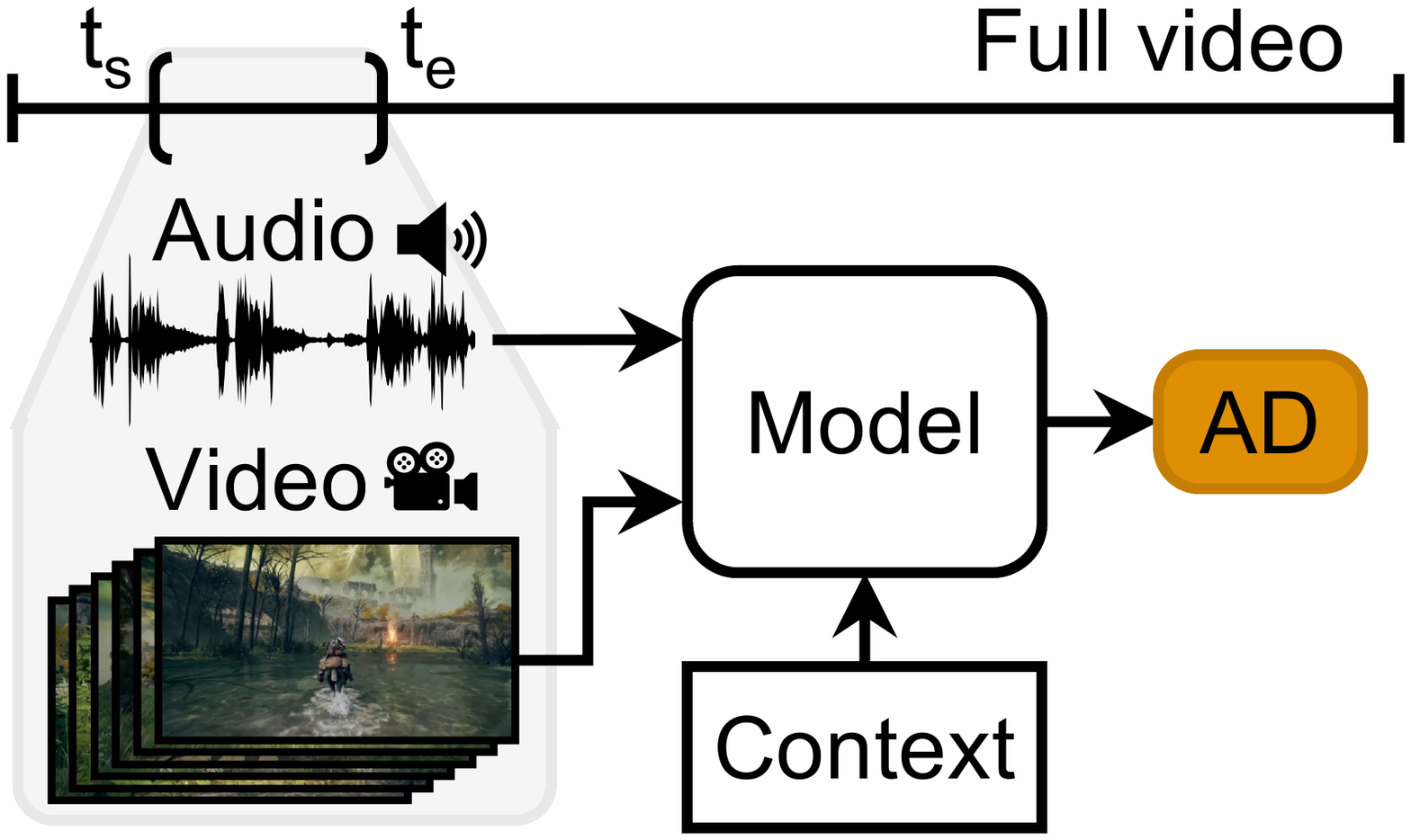}
        \caption{Segment-level task.}
        \label{fig:segment}
    \end{subfigure}%
    \hfill
    \begin{subfigure}[t]{0.32\textwidth}
        \centering
        \includegraphics[width=\linewidth]{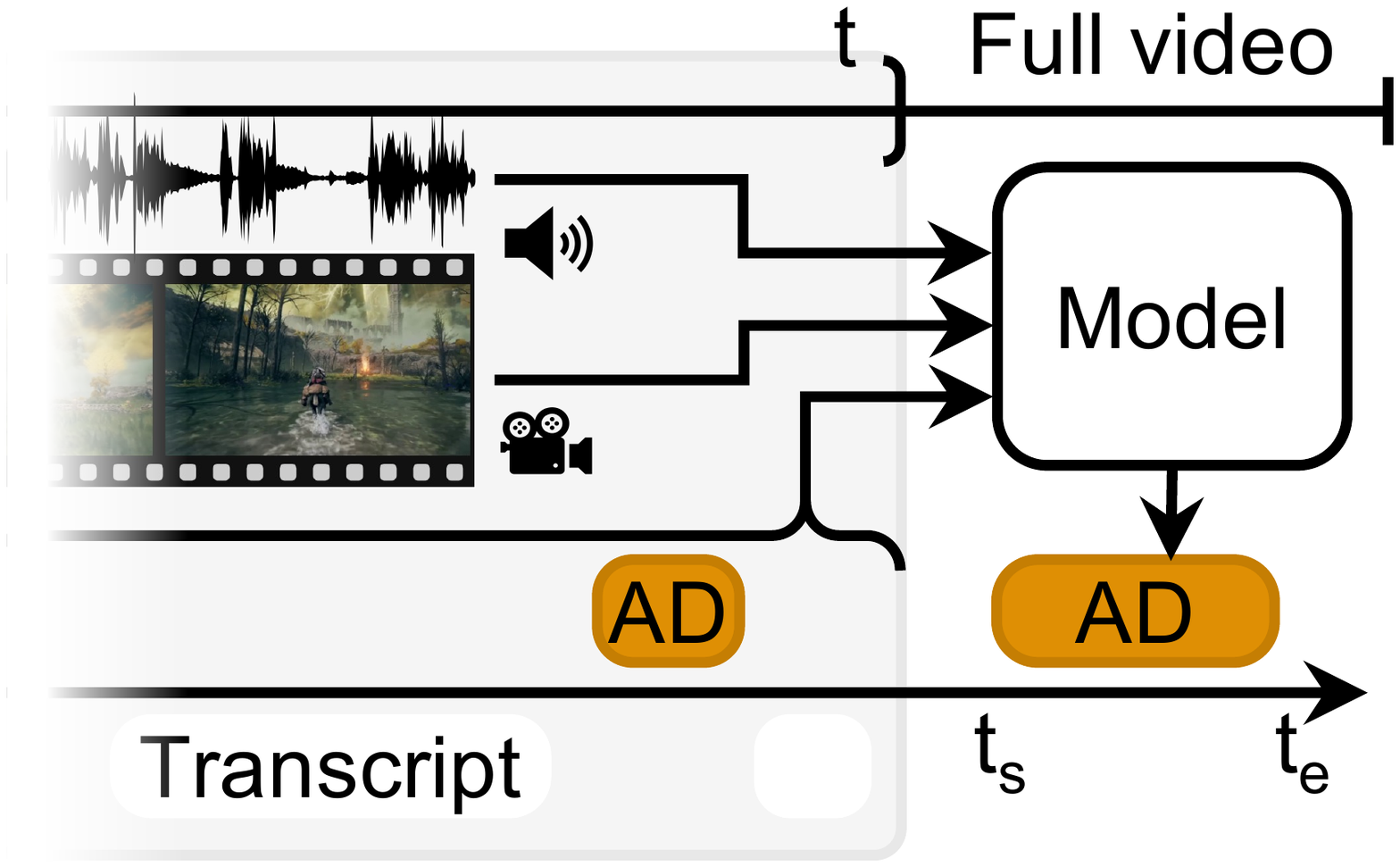}
        \caption{Streaming task.}
        \label{fig:streaming}
    \end{subfigure}%
    \hfill
    \begin{subfigure}[t]{0.32\textwidth}
        \centering
        \includegraphics[width=\linewidth]{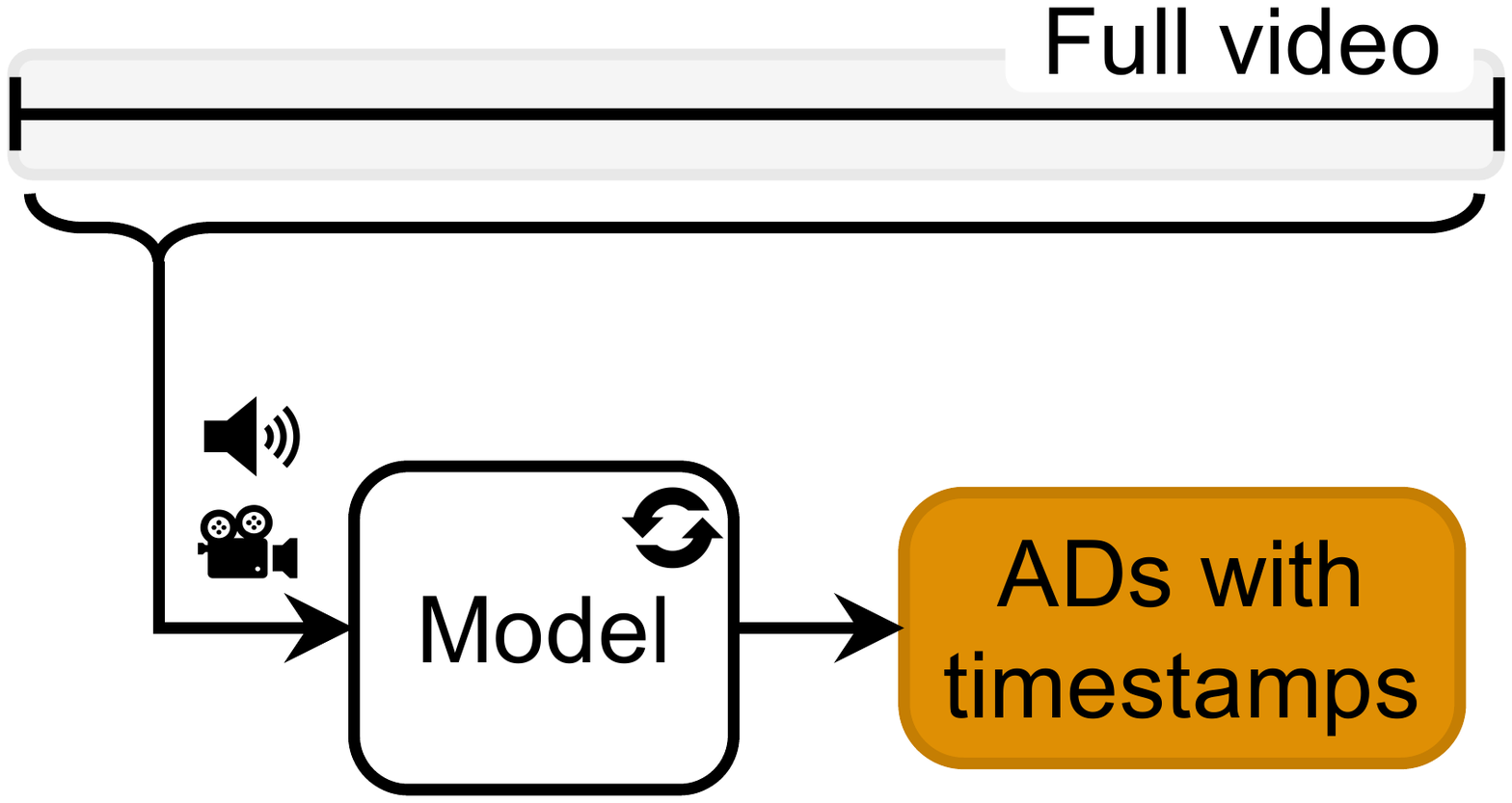}
        \caption{Document-level task.}
        \label{fig:document}
    \end{subfigure}
    \caption{\textbf{Tasks on \audiodescbench{}.} The three tasks differ in input context and the granularity at which ADs are produced: (\subref{fig:segment})~generate an AD for one segment given ground-truth timestamps \(t_s, t_e\) and curated context (e.g., characters, screenplay); (\subref{fig:streaming})~generate the next AD conditioned on video, audio, transcript, and previously generated ADs up to cursor \(t\) advanced by stride \(\delta{}\); (\subref{fig:document})~generate the complete AD transcript for the full video in multiple passes.}
    \label{fig:tasks}
\end{figure}

In this work, we introduce the streaming audio description (\audiodescbench{}) benchmark for full-video AD generation on long-form videos from diverse genres (see \cref{fig:stats}), enabling systematic, automatic evaluation.
We reformulate AD generation as three distinct tasks: the segment-level task, extracting and describing video clips with known ground-truth timestamps and additional metadata; the streaming task, producing AD sentences with timestamps in real time from a sliding window; and the document-level task, generating the ADs for the entire video in multiple passes.
The streaming and document-level tasks move beyond the local, segment-bounded perspective toward long video understanding~\citep{song2024moviechat,ren2024timechat}.
We address the tasks jointly with a zero-shot baseline, \audiodesczeromodel{}, and a fine-tuned baseline, \audiodescmodel{}, which emit AD sentences and timestamps for full videos.

The main contributions of this paper is the first end-to-end streaming AD generation approach for full-length videos, complementing the offline, multi-stage full-video pipelines of \citet{wang2021toward} and \citet{lee2025now}, paired with \audiodescbench{}, the first benchmark of full-length videos of diverse genres with manually verified annotations.
Concretely, our contributions are:
\vspace{-5.5pt}
\begin{itemize}[noitemsep, topsep=0pt, parsep=0pt, leftmargin=1em]
\item \audiodescbench{}, an AD benchmark distinct from prior work in that it features:
    \begin{itemize}[noitemsep, topsep=0pt, parsep=0pt, leftmargin=1em, label=\(\circ\)]
    \item the streaming and document-level tasks to evaluate full video AD generation performance for the first time alongside the established segment-level task,
    \item diverse video genres of different length and AD densities, and
    \item manually verified transcripts and AD sequences;
    \end{itemize}
\item fine-tuned and zero-shot baselines built on Phi-4-mm~\citep{abouelenin2025phi} and Qwen-3.5~\citep{qwen3.5} (\(<\)5B parameters) that approach AD generation as dense video captioning streaming, processing full videos with a sliding window without ground-truth timestamps, where our \audiodescmodel{}:
    \begin{itemize}[noitemsep, topsep=0pt, parsep=0pt, leftmargin=1em, label=\(\circ\)]
    \item matches or exceeds prior methods on the segment-level tasks CMD-AD, MAD-Eval, and \audiodescbench{},
    \item generates coherent full-video AD without ground-truth timestamps on \audiodescbench{}, placing descriptions where the audio leaves room,
    \item enables faster-than-real-time streaming, using \(<\)5B parameters, and
    \item improves streaming localization and reduces redundancy via audio.
    \end{itemize}
\end{itemize}

\section{Related Work}
\citet{lakritz2006semi} leverage screenplays to semi-automatically generate AD, the earliest approach to the task.
More recently, new AD datasets and LLMs have enabled fine-tuning and zero-shot prompting to address single-sentence AD generation.

\textbf{Audio description datasets.}
M-VAD~\citep{torabi2015using} and MPII-MD~\citep{rohrbach2015dataset} were among the first AD datasets, providing 59k and 68k aligned AD sentences from movies, respectively. 
Both datasets were later combined to LSMDC~\citep{rohrbach2017movie}.
MAD~\citep{soldan2022MAD} scales this effort to 384k AD sentences from 650 movies.
MAD-v2~\citep{han2023autoad} is a cleaned revision with 264k sentences across 488 movies with a 10-movie evaluation split called MAD-Eval.
MAD-Eval is a standard evaluation benchmark for segment-level methods~\citep{han2023autoad, han2023autoad2, han2024autoad3, xie2024autoadzero, fang2025distinctad, wang2025uniad, xie2025shotbyshot}.
These datasets are short video clips cropped around the vision-aligned AD events designed for retrieval.
CMD-AD~\citep{bain2020CMD, han2024autoad3} provides 101k AD sentences from 1,432 YouTube-sourced movie clips paired with a character identity database~\citep{han2023autoad2}.
The AudioVault community\footnote{https://audiovault.net/} contributes a large text-only AD corpus~\citep{han2023autoad}.
TV-AD~\citep{xie2024autoadzero} focuses on television shows with 31k training and 3k evaluation sentences.
ADAQ~\citep{kala2025adaq} is a benchmark with LLM-generated visual and narrative question-answer pairs based on CMD-AD.
Our \audiodescbench{} benchmark is the first purpose-built benchmark for full-video AD generation with temporal alignment covering full-length videos across diverse genres.

\textbf{Audio description generation.}
Segment-level AD generation produces one description per video segment given the ground-truth timestamps.
There are fine-tuned~\citep{han2023autoad, han2023autoad2, han2024autoad3, lin2024learning, wang2025uniad, fang2025distinctad} and training-free approaches~\citep{xie2024autoadzero, chu2024llm, xie2025shotbyshot, park2025narrad, khandelwal2025coherentad, zhang2024mmnarrator} to the segment-level task.
Training-free pipelines chain pre-trained vision models and LLMs, combining face detection, character identification, video captioning, and prompting~\citep{xie2024autoadzero, chu2024llm, xie2025shotbyshot, park2025narrad, khandelwal2025coherentad, zhang2024mmnarrator}.
Since the segment-level task only provides a partial view of the full problem, the main focus is on enriching metadata.
Some approaches inject character identities directly into the prompt~\citep{han2023autoad2,han2024autoad3,fang2025distinctad}, while others detect character faces and mark them on the frame~\citep{xie2024autoadzero, chu2024llm, xie2025shotbyshot}, optionally with character relevance classification~\citep{wang2025uniad}.
Additional context include temporal context such as previous and future frames~\citep{wang2025uniad, yang2025story, fang2025distinctad}, screenplays~\citep{lakritz2006semi,park2025narrad}, dialogue~\citep{zhang2024mmnarrator}, and generated AD~\citep{zhang2024mmnarrator}.
Domain-specific approaches focus on movies in general~\citep{xie2025shotbyshot, yang2025story}, animated movies~\citep{gui2025character}, soccer~\citep{chaudhary2025mcad, rao2024matchtime}, or basketball~\citep{xi2025simple}.
A mostly unaddressed challenge is to generate AD for full videos.
\citet{wang2021toward} proposed a pipeline that includes localization, candidate generation, and cost minimization to find a sequence with minimal irrelevance, diversity, and perplexity cost.
Related work classifies AD gaps~\citep{han2023autoad2, lee2025now}, fits AD into speech gaps~\citep{park2025narrad}, reduces repetition~\citep{fang2025distinctad, khandelwal2025coherentad}, and incorporates prior predictions~\citep{zhang2024mmnarrator}.

\textbf{Dense video captioning.}
Dense video captioning generates timestamped visual descriptions~\citep{krishna2017dense}.
Vid2Seq~\citep{yang2023vid2seq} applies dense video captioning to the full video, outputting a sequence of captions with temporal grounding.
\citet{zhou2024streaming} processes videos in a streaming fashion by building a memory via online clustering of frame embeddings, generating captions at decoding points.
In contrast, AD events need to be story-relevant, are limited in the number of words, must not override relevant audio, and are not perfectly aligned to the visual event.

\textbf{Audio descriptions evaluation.}
Segment-level AD is typically evaluated with image captioning metrics, including CIDEr~\citep{vedantam2015cider}, METEOR~\citep{banerjee2005meteor}, and ROUGE-L~\citep{lin2004rouge}.
Specialized AD metrics include Recall@\(k\)/\(N\)~\citep{han2023autoad2}, LLM-AD-eval~\citep{han2024autoad3}, and CRITIC~\citep{han2024autoad3}.
Recall@\(k\)/\(N\) checks whether the predicted AD falls in the top-\(k\) of a window of size \(N\).
LLM-AD-eval uses an LLM to rate the predicted AD against the ground truth AD.
It is the most popular of the other AD-focused LLM-based metrics~\citep{zhang2024mmnarrator, kala2025adaq}.
CRITIC rates the correct appearance of character names in AD.
More recently, \citet{xie2025shotbyshot} introduce ActionScore to measure action coverage in AD.
\citet{lee2025now} evaluate localization with recall, precision, and F1 at tIoU \(>\) 0.1, and AD generation with image captioning metrics on temporally closest prediction–reference pairs.
Dense video captioning averages such metrics over multiple tIoU thresholds~\citep{krishna2017dense}, with SODA measuring correct temporal order~\citep{fujita2020soda}.

\section{\audiodescbench{} Benchmark}
\begin{figure}[t]
    \centering
    \includegraphics[width=\linewidth]{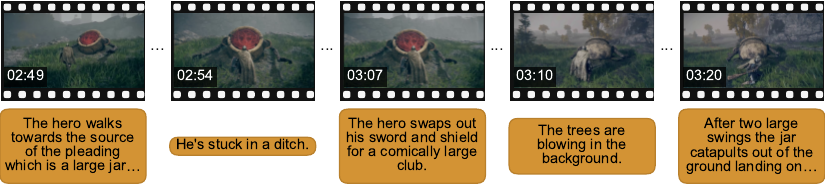}
    \caption{\textbf{Example from \audiodescbench{}.} An example sequence from \emph{Elden Ring---Gameplay Preview} (\href{https://youtu.be/JldMvQMO_5U}{\nolinkurl{youtu.be/JldMvQMO_5U}}) and its audio-described version (\href{https://youtu.be/neQPEIAqpQ8}{\nolinkurl{youtu.be/neQPEIAqpQ8}}).}
    \label{fig:example}
\end{figure}

Current AD datasets are either focused on movies, e.g., MAD-Eval~\citep{soldan2022MAD, han2023autoad} and CMD-AD~\citep{bain2020CMD, han2024autoad3}, or on television shows, e.g., TV-AD~\citep{xie2024autoadzero}.
Annotations are not intended directly for AD generation, e.g., MAD-Eval is intended for retrieval~\citep{soldan2022MAD}, or are not manually verified, e.g., CMD-AD uses an automatic annotation pipeline~\citep{han2024autoad3}.
Furthermore, the existing datasets consist of short clips with pre-defined segment boundaries.
We introduce the Streaming Audio Description (\audiodescbench{}) benchmark for evaluating AD generation for full-length videos in the wild.
The benchmark is visualized in \cref{fig:stats,fig:tasks,fig:example}.
Details about dataset creation and reproducibility can be found in \cref{sec:appendix-benchmark}.

\textbf{Task definition.}
We propose three tasks on \audiodescbench{} visualized in \cref{fig:tasks}, listed below:
\vspace{-4pt}
\begin{enumerate}[noitemsep, topsep=0pt, parsep=0pt, leftmargin=1em]
\item{
\textbf{Segment-level task.} 
Segment-level AD generation is the standard formulation~\citep{han2023autoad}: given a short video segment with known boundaries, generate a single AD sentence describing the visual content.
Additional contextual information, such as character databases or screenplays, may be used as input because the task is inherently ambiguous.
}
\item{
\textbf{Streaming task.}\label{sec:streaming_task}
Streaming AD generation moves a temporal cursor forward to generate AD sentences with temporal localization in real time for an entire video.
The system must infer characters, relations, and key objects from the generated context, the current video, and its transcript, without access to ground-truth timestamps or other metadata.
We treat the speech transcript as part of the input signal rather than as auxiliary metadata, since transcripts are routinely available from on-platform subtitles or off-the-shelf ASR.
}
\item{
\textbf{Document-level task.}
Document-level AD generation is the offline, full-video AD generation task. 
It explicitly incorporates derivatives of the entire video to iteratively generate the complete ADs, in multiple passes, without relying on external metadata.
}
\end{enumerate}

\textbf{Statistics.}
\audiodescbench{} comprises 33 videos totaling 22.3 hours, spanning durations from 1 minute to 5.8 hours (median 21.5 minutes) across five genres: short films, documentaries, video games, movies, and performances.
The videos contain 9{,}131 manually verified AD events (6.8 hours, 84k words), with events averaging 2.7 s and 9.3 words, and per-video AD density ranging from 0.1 to 0.9.
This makes \audiodescbench{} the largest AD evaluation set by AD events (vs. 7{,}316 for CMD-AD test and 6{,}520 for MAD-Eval) and the only one composed of full-length videos.
The dataset has one overlap with an existing dataset based on the IMDb ID~\footnote{\href{https://www.imdb.com/}{imdb.com}}, that is the ID \texttt{tt0031397} also included in the AudioVault train dataset.

\begin{algorithm}[t]
\caption{AD Streaming Inference}\label{alg:streaming}
\KwIn{transcript segments \(\mathcal{T}\), video \(\mathcal{V}\), text window \(w_c\), video window \(w_v\), duration \(D\), stride \(\delta{}\)}
\KwOut{predicted AD segments \(\mathcal{A}\)}
\(\mathcal{A} \leftarrow \emptyset\)\;
\(t \leftarrow 0\)\;
\While{\(t < D\)}{
 \(t \mathrel{+}= \delta{}\)\;
 \lIf{\(\neg\,\textsc{HasGap}(\mathcal{T},\; t - \delta,\; t)\)}{\textbf{continue}}
 \(\textrm{prediction} \leftarrow \textsc{Generate}(\mathcal{V}[t - w_v,\; t],\; \mathcal{T}[t - w_c,\; t],\; \mathcal{A}[t - w_c,\; t])\)\;
\lIf{\(\textrm{prediction}\) \textrm{invalid} \(\lor\) \(\textrm{prediction} = \textsc{Wait}\)}{\textbf{continue}}
\(\mathcal{A} \leftarrow \mathcal{A} \cup \{\textrm{prediction}\}\), \(t \leftarrow \max(t,\; t - w_v + \textrm{prediction}.t_e)\)\;
}
\end{algorithm}

\section{Audio Description Streaming}
Current AD generation methods assume the availability of ground-truth context at inference time, e.g., segment boundaries, character information, or screenplays~\citep{han2023autoad, han2023autoad2, han2024autoad3, xie2025shotbyshot}.
Such metadata and especially AD event timestamps are rarely available for arbitrary videos.
Existing approaches aiming for the full-video setting generally fall into the document-level category, requiring iterative refinement~\citep{wang2021toward, park2025narrad, khandelwal2025coherentad, lee2025now}.
Generating ADs for full videos in real time and without prior context would be a valuable tool to make, for example, online video, live events, or interactive media such as video games~\citep{larreina2024games} accessible.
We propose two baselines for joint AD segment-level generation and streaming: a zero-shot approach building on AutoAD-Zero~\citep{xie2024autoadzero}, \audiodesczeromodel{}, and a fine-tuned approach, \audiodescmodel{}.
Inspired by whisper~\citep{radford2023robust} and long video understanding methods~\citep{zhou2024streaming, song2024moviechat, ren2024timechat}, both baselines use a sliding window to process the full video.
The algorithm finds speech gaps and iteratively fills them with AD sentences, advancing the window via estimated timestamps.

\subsection{\audiodesczeromodel}\label{sec:audiodesczeromodel}
\audiodesczeromodel{}, is built on AutoAD-Zero~\citep{xie2024autoadzero}, with a segment-level and a streaming path.
The segment-level path follows the original AutoAD-Zero approach: sequential execution of (i) a face detection and character identification stage, (ii) a video captioning stage, and (iii) an AD generation stage.
If character identities are available, they are added to the prompt and displayed with colored bounding boxes.
We adapt AutoAD-Zero to streaming by extending the AD generation stage to consume previously generated ADs and the transcript within a context window \(w_c = 120\,\text{s}\).
The video captioning stage uses a visual window \(w_v = \delta\) equal to the stride, so it only describes new visuals.
Character information is not used for streaming.
Before AD generation, we add a decision stage that predicts whether the current window should emit an AD, conditioned on prior context and the current video caption.
Assuming a fixed speech rate of 150 words per minute, we estimate the AD's spoken duration and center it within the visual window to derive its timestamps.
Prompts are documented in \cref{sec:appendix-zeroshot-prompts}.
We use Qwen-3.5-4B~\citep{qwen3.5} for all stages in \audiodesczeromodel{}.

\subsection{\audiodescmodel}

Fine-tuned models achieve state-of-the-art results on segment-level AD generation~\citep{han2023autoad, han2023autoad2, han2024autoad3, fang2025distinctad, wang2025uniad}.
However, as they rely on ground-truth segment boundaries, they cannot be applied directly to the streaming setting.
We propose \audiodescmodel{} that is jointly trained on segment-level task and streaming task to predict AD sentences with timestamps.
We train the projection layers of a pre-trained MLLM for vision and, if supported, audio.
We use LoRA~\citep{hu2022lora} to adapt the language model layers.

For the segment-level task, zero-shot approaches typically pass the AD duration~\citep{xie2024autoadzero} or an AD word count estimate derived from dataset-specific speech rates~\citep{xie2025shotbyshot}.
We sample 8 evenly-spaced frames and audio from a padded segment \([t_s - \delta_p,\; t_e + \delta_p]\), with \(\delta_p = 2\,\text{s}\) for both training and inference.
For MAD-Eval, we set \(\delta_p = 0\,\text{s}\) since the released clips contain no surrounding context.
By fixing \(\delta_p\), the model can infer the duration of the AD event and, consequently, the number of words required to describe it.
We overlay character information in the same way as for AutoAD-Zero~\citep{xie2024autoadzero} (\cref{sec:audiodesczeromodel}).

For the streaming task, we use a fixed visual window of \(w_v = 8\,\text{s}\).
The transcript and previous ADs are included from the context window \(w_c = 120\,\text{s}\).
Timestamps are encoded inline as tokenized text inserted at the start and end of each element in the sequence (\texttt{(t=0.1)} for Phi-4-mm and \texttt{<0.1 seconds>} for Qwen-3.5).
Character identities are not provided.
Each prediction is prefixed with \texttt{AD} or \texttt{WAIT}.
Wait signals serve three purposes: deferring when dialogue occupies the gap, preserving intentional audio (e.g., music or sound effects), and pausing until sufficient visual context has accumulated.
We create streaming data by moving a window with a random stride over each video.
An unseen AD event becomes a sample, while previous events and transcripts are used for context.
Samples are labeled as \texttt{WAIT} if the window contains just dialogue, if it contains no ADs, or if it has a partial AD event (i.e., mostly outside the window).

We fine-tune the multimodal model Phi-4-mm~\citep{abouelenin2025phi} and the vision-language model Qwen-3.5~\citep{qwen3.5}.
Both models are extensively pre-trained on web-scale data.
For Phi-4-mm~\citep{abouelenin2025phi}, we reduce the model size to 4.8B parameters by merging the pre-trained vision LoRA with the main weights.

\subsection{Streaming Model Deployment}
At inference, \audiodesczeromodel{} and \audiodescmodel{} process a full-length video by sweeping a backwards-facing sliding window from start to end and emitting AD sentences as it advances (\cref{alg:streaming}).
The window position is tracked by a cursor \(t\) that moves through the full video duration.
For each iteration, the method advances \(t\) by stride \(\delta\) until transcript \(\mathcal{T}\) has a sufficient gap (\(\geq 0.5\)s).
The model takes as input the visual window \(w_v\) and text context window \(w_c\) (\(\mathcal{T}\) and generated ADs \(\mathcal{A})\) to generate the next AD.
If generation fails or returns a \texttt{WAIT} signal, the loop continues.
Otherwise, if the predicted end time \(t_e\) exceeds the current \(t\), we set \(t = t_e\).

\section{Experiments}\label{sec:experiments}
Our experiments support our claims that our approach
(i) matches or exceeds prior and baseline methods on CMD-AD, MAD-Eval, and \audiodescbench{},
(ii) generates coherent full-video AD without ground-truth timestamps,
(iii) is more efficient than our zero-shot baseline, enabling real-time streaming on an A100 GPU, and 
(iv) improves streaming localization and reduces redundancy via audio.

\subsection{Experimental Setup}

\textbf{Datasets.}
We train on CMD-AD~\citep{bain2020CMD, han2024autoad3}, using the 7{,}456 of 8{,}339 clips available at publication time, and hold out 25 movies with 147 clips for validation (\cref{sec:cmd-ad-val-split}).
We evaluate segment-level performance on CMD-AD (551 of 591 test clips, 6{,}740 AD events), MAD-Eval~\citep{soldan2022MAD,han2023autoad} (10 movies, 6.5K AD events), and \audiodescbench{}.
We verify in \cref{sec:appendix-incomplete-cmdad} that the incomplete test set does not change the relative ranking of methods.
For CMD-AD and MAD-Eval we use the character database from \citet{han2023autoad2}.
Streaming is evaluated on \audiodescbench{}, with videos split into 30-minute chunks.

\textbf{Training.}
We use AdamW~\citep{loshchilov2017decoupled} with cross-entropy loss, training for 2 epochs with early stopping following \citet{han2024autoad3}.
We use LoRA with \(r{=}16\) and \(\alpha{=}32\), batch size 8, and tune the learning rate per backbone with Optuna~\citep{akiba2019optuna} (\(9.2\mathrm{e}{-5}\) for Phi-4-mm, \(4.3\mathrm{e}{-5}\) for Qwen-3.5).
Of the sample budget, 25\,\% is allocated to \texttt{WAIT} signals and the remainder split equally between segment-level and streaming samples.
Full hyperparameters and hardware setup are reported in \cref{sec:appendix-detailed-setup}.

\textbf{Inference.}
We use greedy decoding, with a presence penalty~\citep{keskar2019ctrl} of 1.5 for streaming to reduce repetitions.
For segment-level evaluation, the model receives only character names and face detections, matching prior work.
For streaming, each window is conditioned on previously generated ADs and the ground-truth transcript overlapping the text context window.
We use the verified transcript to isolate AD generation from ASR error.

\textbf{Metrics.}
For segment-level evaluation we report CIDEr~\citep{vedantam2015cider}, Recall@\(k\)/\(N\)~\citep{han2023autoad2}, CRITIC~\citep{han2024autoad3}, and LLM-AD-Eval~\citep{han2024autoad3}.
For streaming, we group metrics into four categories.
\emph{Quality} adopts standard dense video captioning metrics~\citep{krishna2017dense, yang2023vid2seq, zhou2024streaming}: SODA~\citep{fujita2020soda}, CIDEr, and METEOR~\citep{banerjee2005meteor}, averaged over tIoU thresholds 0.1, 0.3, and 0.5.
\emph{Temporal localization} reports Recall and Precision over the same tIoU-matched pairs.
\emph{Diagnostics} target streaming-specific failures: repetition (REP) measures the fraction of consecutive AD pairs that are exact duplicates (METEOR\,\(=1\)), and overlap (OL) measures the fraction of predicted AD duration overlapping transcribed speech.
\emph{Efficiency} reports the real time factor (RTF) and lag (LAG), averaged over AD-emitting windows.
Metric definitions and a reproducibility analysis are given in \cref{sec:appendix-metrics}.

\begin{table}[!t]
 \centering
 \caption{Segment-level evaluation on CMD-AD, MAD-Eval, and \audiodescbench{}.}
 \label{tab:segment}
 \footnotesize
 \setlength{\tabcolsep}{1pt}
 \begin{tabular}{clcccccccc}
  \toprule
  & & \multicolumn{4}{c}{CMD-AD} & \multicolumn{2}{c}{MAD-Eval} & \multicolumn{2}{c}{\audiodescbench{}} \\
  \cmidrule(lr){3-6} \cmidrule(lr){7-8} \cmidrule(lr){9-10}
  & Method & CIDEr\up{} & R@1/5\up{} & CRITIC\up{} & LLM-AD-Eval\(^{\ddagger}\)\up{} & CIDEr\up{} & R@5/16\up{} & CIDEr\up{} & R@5/16\up{} \\
  \midrule
  \multirow{8}{*}{\rotatebox[origin=c]{90}{Fine-tuned}} & AutoAD-II~\citep{han2023autoad2} & 13.5 & 26.1 & 8.2 & 2.08 | \phantom{0.00} & 19.5 & 51.3 &  &  \\
   & AutoAD-III~\citep{han2024autoad3} & 25.0 & 31.2 & 32.7 & 2.89 | 2.01 & 24.0 & 52.8 &  &  \\
   & MovieSeq~\citep{lin2024learning} &  &  &  &  & 24.4 & 51.6 &  &  \\
   & DistinctAD~\citep{fang2025distinctad} & 22.7 & 33.0 &  & 2.88 | 2.03 & 27.3 & 56.0 &  &  \\
   & UniAD~\citep{wang2025uniad} &  &  &  &  & \textbf{28.2} & 54.9 &  &  \\
  \cmidrule(lr){2-10}
   & AutoAD-III~\citep{han2024autoad3}\(^{\dagger}\) & 21.6 & 30.2 & 26.7 & 3.26 | 2.35 & 21.7 & 51.1 & 17.5 & 45.2 \\
   & \audiodescmodel{} (Phi-4-mm, ours) & 29.8 & 35.6 & 29.6 & 3.35 | 2.51 & 19.4 & 49.9 & 32.3 & 52.6 \\
   & \audiodescmodel{} (Qwen-3.5, ours) & \textbf{36.3} & \textbf{38.0} & 31.8 & 3.44 | 2.74 & 24.9 & 54.8 & \textbf{51.0} & \textbf{57.9} \\
  \midrule
  \multirow{8}{*}{\rotatebox[origin=c]{90}{Zero-shot}} & MM-Narrator\(^{\star}\)~\citep{zhang2024mmnarrator} &  &  &  &  & 13.9 & 49.0 &  &  \\
   & LLM-AD\(^{\star}\)~\citep{chu2024llm} &  &  &  &  & 20.5 &  &  &  \\
   & AutoAD-Zero~\citep{xie2024autoadzero} & 17.7 & 26.9 & 43.7 & 2.83 | 1.96 & 22.4 & 47.0 &  &  \\
   & Shot-by-shot~\citep{xie2025shotbyshot} & 26.3 & 33.0 & 47.8 & 3.15 | 2.42 & 25.0 & 50.6 &  &  \\
   & StoryContext-AD~\citep{yang2025story} &  &  &  &  & 26.0 & 50.3 &  &  \\
   & NarrAD\(^{\star}\)~\citep{park2025narrad} &  &  &  &  & 26.4 & 54.0 &  &  \\
   & Shot-by-shot\(^{\star}\)~\citep{xie2025shotbyshot} & 26.1 & 36.5 & \textbf{49.1} & 3.17 | 2.66 & 26.9 & 56.4 &  &  \\
  \cmidrule(lr){2-10}
  & \audiodesczeromodel{}~\citep{xie2024autoadzero} (Qwen-3.5, ours) & 18.8 & 29.8 & 44.0 & \textbf{3.48} | \textbf{2.86} & 22.0 & \textbf{57.1} & 12.6 & 45.8 \\
  \bottomrule
  \multicolumn{10}{l}{\begin{minipage}{0.9\linewidth}
   \scriptsize \(\dagger\)~denotes our reproduction. \(\ddagger\)~LLM-AD-Eval~\citep{han2024autoad3} reported as Llama-2-7B~\citep{touvron2023llama}~\(\vert\)~Llama-3-8B~\citep{grattafiori2024llama} as in~\citep{xie2024autoadzero}. \(\star\)~are proprietary.
  \end{minipage}} \\
 \end{tabular}
\end{table}

\subsection{Segment-level Evaluation}\label{sec:segment-level-eval}
We compare against prior zero-shot and fine-tuned segment-level methods to support claim~(i) that our approach matches or exceeds their AD quality.
The results are reported in \cref{tab:segment}.

\textbf{Validating the zero-shot and fine-tuned baselines.}
Built on AutoAD-Zero~\citep{xie2024autoadzero}, \audiodesczeromodel{} achieves a small improvement (+1.1\,CIDEr) using only Qwen-3.5 4B, despite using less than a third of the parameters (AutoAD-Zero uses VideoLLaMA2-7B~\citep{cheng2024videollama} and LLaMA3-8B~\citep{grattafiori2024llama}).
Next, we re-ran AutoAD-III~\citep{han2024autoad3}.
Our reproduced numbers for CMD-AD are lower than the published ones (21.6\,\(<\)\,25.0\,CIDEr), likely due to differences in the character database and model inference setup.
On \audiodescbench{}, AutoAD-III performs better than \audiodesczeromodel{} (+4.9\,CIDEr), while R@5/16 is close, indicating a better use of AD vocabulary.

\textbf{Comparing results on CMD-AD and \audiodescbench{}.}
Our Qwen-3.5 and Phi-4-mm \audiodescmodel{} are the smallest models in the comparison.
\audiodescmodel{} sets the state of the art on CMD-AD with improvements of +10.0\,CIDEr over Shot-by-shot~\citep{xie2025shotbyshot} and +11.3\,CIDEr over AutoAD-III~\citep{han2024autoad3}, improving over fine-tuned and proprietary models.
Similarly, R@1/5 is the highest with 38.3, 1.8 points higher than Shot-by-shot.
Previous fine-tuned models rely on extensive pre-pretraining on AD-related data~\citep{han2023autoad2,han2024autoad3,wang2025uniad}.
Instead \audiodescmodel{} is fine-tuned only on CMD-AD and relies on broad vision-language pre-training, extended vision context, and temporal cues.
The CRITIC metric is dominated by zero-shot approaches, while for fine-tuned models AutoAD-III~\citep{han2024autoad3} has the highest score.
We hypothesize that zero-shot approaches are more verbose and name characters more frequently, though whether this benefits AD quality remains an open question for future work.
On \audiodescbench{}, our Qwen-3.5 \audiodescmodel{} sets the highest score with 51.0\,CIDEr, ahead of Phi-4-mm \audiodescmodel{} (+18.7\,CIDEr), AutoAD-III (+33.5\,CIDEr), and \audiodesczeromodel{} (+38.4\,CIDEr).

\textbf{Investigating the MAD-Eval domain-gap.}
Temporal cues~\citep{wang2025uniad,xie2024autoadzero,xie2025shotbyshot} and extended vision context~\citep{wang2025uniad,xie2025shotbyshot,fang2025distinctad} are commonly used.
However, the impact of temporal cues is rarely discussed.
We observe that the CMD-AD performance does not transfer to MAD-Eval.
Our \audiodescmodel{} achieves a lower score than UniAD~\citep{wang2025uniad} (-3.3\,\text{CIDEr}) and DistinctAD~\citep{fang2025distinctad} (-2.4\,\text{CIDEr}).
Notably, the R@5/16 differs by only 0.1 for UniAD, suggesting similar semantics.
The CIDEr metric is sensitive to sentence length~\citep{vedantam2015cider}, requiring temporal cues to estimate the word count in the segment-level task.
Furthermore, MAD is intended for video clip retrieval, aligning and expanding AD events to visual events~\citep{soldan2022MAD}.
UniAD~\citep{wang2025uniad} and DistinctAD~\citep{fang2025distinctad} are fine-tuned on MAD-v2~\citep{soldan2022MAD,han2023autoad} learning the MAD-specific temporal cues.
Conversely, DistinctAD does not generalize well to CMD-AD.
Shot-by-shot addresses this gap by providing a dataset-specific speech rate factor to the prompt~\citep{xie2025shotbyshot}.
We train a variant of \audiodescmodel{} that uses a random padding for segments \(\delta_p \sim \mathcal{U}(0, 2)\,\text{s}\) instead of \(\delta_p = 2\,\text{s}\).
This model achieves 31.1 CIDEr on CMD-AD and 31.9 on MAD-Eval.
Although this model achieves state-of-the-art results for CMD-AD and MAD-Eval, we opted for the model with temporal cues because it preserves the AD temporal alignment. 
Further discussion of this trade-off will be deferred to future work.

Overall, these results support claim~(i): \audiodescmodel{} matches or exceeds prior segment-level methods using fewer parameters and only the CMD-AD training data.
At the same time, the gap between CMD-AD and MAD-Eval justifies the need for \audiodescbench{}.

\begin{table}[t]
 \centering
 \caption{
  Streaming evaluation on \audiodescbench{}.
 }
 \label{tab:streaming}
 \footnotesize
 \setlength{\tabcolsep}{3pt}
 \begin{tabular}{lccccccccc}
  \toprule
  & \multicolumn{3}{c}{Quality} & \multicolumn{2}{c}{Localization} & \multicolumn{2}{c}{Diagnostics} & \multicolumn{2}{c}{Efficiency} \\
  \cmidrule(lr){2-4} \cmidrule(lr){5-6} \cmidrule(lr){7-8} \cmidrule(lr){9-10}
  Method & SODA\up{} & CIDEr\up{} & METEOR\up{} & R\up{} & P\up{} & REP\down{} & OL\down{} & RTF\down{} & LAG\down{} \\
  \midrule
  \audiodescmodel{} (Phi-4-mm, ours) & 1.6 & 16.3 & 5.3 & 57.3 & 55.9 & 19.5 & 2.0 & \textbf{0.47} & \textbf{4.0} \\
  \audiodescmodel{} (Qwen-3.5, ours) & \textbf{2.4} & \textbf{34.0} & \textbf{7.5} & \textbf{59.5} & \textbf{57.8} & \textbf{0.7} & \textbf{1.5} & 0.52 & 4.2 \\
  \midrule
  \audiodesczeromodel{} (Qwen-3.5, ours) & 1.1 & 9.5 & 5.0 & 31.9 & 52.7 & 4.1 & 8.2 & 2.92 & 5.0 \\
  \bottomrule
 \end{tabular}
\end{table}

\subsection{Streaming Evaluation}
We evaluate the full streaming pipeline on \audiodescbench{} to support claim~(ii) that our approach generates coherent AD across a full video without ground-truth timestamps.
The results are reported in \cref{tab:streaming}.

\textbf{Comparing results.}
The fine-tuned Qwen-3.5 and Phi-4-mm \audiodescmodel{} perform better than \audiodesczeromodel{} (+1.3\,SODA and +24.5\,CIDEr).
Qualitative examples (see \cref{sec:appendix-qualitative}) suggest that fine-tuned models tend to be more verbose, whereas zero-shot models tend to write longer sentences and pause more frequently.
Despite its additional audio input, Phi-4-mm scores lower than Qwen-3.5 (-1.0\,SODA, -17.7\,CIDEr), likely because the more recent Qwen-3.5 has stronger video understanding.
Qwen-3.5 also shows better temporal localization than Phi-4-mm (+2.2\,Recall and +1.9\,Precision) and \audiodesczeromodel{} (+27.6\,Recall and +5.1\,Precision).
Regarding diagnostics, we see that \audiodesczeromodel{} has a large overlap with speech while fine-tuned models successfully reduce the overlap.
Both \audiodesczeromodel{} and Phi-4-mm \audiodescmodel{} tend to repeat sentences, Phi-4-mm shows a severe collapse though.
This problem is also studied in text-only LLMs~\citep{su2022contrastive}.
However, existing approaches that avoid repetition~\citep{wang2021toward, fang2025distinctad, khandelwal2025coherentad} do not transfer to the streaming setting or cause high output variability.
Inference strategies like contrastive search can help but increase computational cost~\citep{su2022contrastive}.
Qwen-3.5 is only slightly affected by this problem.

Overall, these results support claim~(ii): \audiodescmodel{} produces coherent AD across a full video without requiring ground-truth timestamps.
The gap to the segment-level upper bound and the streaming-specific failure modes confirm that \audiodescbench{} sets a meaningful and unsolved target for future work.

\subsection{Efficiency}

We measure the inference speed on the streaming task to support claim~(iii) that our approach runs in real time.
The speed is measured by executing our approach on the first 2 minutes of each video in \audiodescbench{}.
The measurements are reported in \cref{tab:streaming}.
Both Qwen-3.5 and Phi-4-mm \audiodescmodel{} run faster than real time, with RTF 0.52 and 0.47 respectively, and a lag of about 4\,s.

\begin{table}[t]
 \centering
 \caption{Phi-4-mm \audiodescmodel{} training ablations on the CMD-AD validation split.}
 \label{tab:ablation-modality}
 \footnotesize
 \setlength{\tabcolsep}{3pt}
 \begin{tabular}{l c c c c c c c c c c c c}
  \toprule
  & \multicolumn{7}{c}{Augmentations \textit{(stepwise removal)}} & & \multicolumn{4}{c}{Modalities} \\
  \cmidrule(lr){2-8} \cmidrule(lr){10-13}
  & \makebox[0pt][l]{\rotatebox[origin=lB]{45}{Baseline}} & \makebox[0pt][l]{\rotatebox[origin=lB]{45}{- segment aug.}} & \makebox[0pt][l]{\rotatebox[origin=lB]{45}{- subtitle aug.}} & \makebox[0pt][l]{\rotatebox[origin=lB]{45}{- face drop}} & \makebox[0pt][l]{\rotatebox[origin=lB]{45}{- context aug.}} & \makebox[0pt][l]{\rotatebox[origin=lB]{45}{- audio aug.}} & \makebox[0pt][l]{\rotatebox[origin=lB]{45}{- frame aug.}} & & \makebox[0pt][l]{\rotatebox[origin=lB]{45}{All modalities}} & \makebox[0pt][l]{\rotatebox[origin=lB]{45}{No audio}} & \makebox[0pt][l]{\rotatebox[origin=lB]{45}{No video}} & \makebox[0pt][l]{\rotatebox[origin=lB]{45}{Context only}} \\
  \midrule
  \multicolumn{13}{l}{\textit{Streaming}} \\
  \midrule
  SODA\up{} & 1.4 & \textcolor[HTML]{D55E00}{1.4} & \textcolor[HTML]{029E73}{1.6} & \textcolor[HTML]{D55E00}{1.5} & \textcolor[HTML]{029E73}{1.5} & \textcolor[HTML]{029E73}{1.6} & \textcolor[HTML]{029E73}{1.6} & & \textbf{1.4} & 1.4 & 0.9 & 0.9 \\
  CIDEr\up{} & 10.9 & \textcolor[HTML]{D55E00}{10.7} & \textcolor[HTML]{D55E00}{10.3} & \textcolor[HTML]{029E73}{10.9} & \textcolor[HTML]{029E73}{11.3} & \textcolor[HTML]{029E73}{12.3} & \textcolor[HTML]{D55E00}{11.1} & & \textbf{10.7} & 9.7 & 5.4 & 5.6 \\
  OL\down{} & 4.6 & \textcolor[HTML]{029E73}{3.9} & \textcolor[HTML]{D55E00}{4.1} & \textcolor[HTML]{D55E00}{4.3} & \textcolor[HTML]{D55E00}{5.1} & \textcolor[HTML]{029E73}{4.4} & \textcolor[HTML]{029E73}{4.3} & & \textbf{3.9} & 5.3 & 5.3 & 6.1 \\
  REP\down{} & 15.7 & \textcolor[HTML]{029E73}{11.5} & \textcolor[HTML]{D55E00}{12.5} & \textcolor[HTML]{029E73}{8.3} & \textcolor[HTML]{D55E00}{19.1} & \textcolor[HTML]{029E73}{16.0} & \textcolor[HTML]{D55E00}{16.3} & & \textbf{11.5} & 16.3 & 49.6 & 39.0 \\
  \midrule
  \multicolumn{13}{l}{\textit{Segment-level}} \\
  \midrule
  CIDEr\up{} & 24.3 & \textcolor[HTML]{029E73}{32.4} & \textcolor[HTML]{029E73}{33.6} & \textcolor[HTML]{029E73}{34.8} & \textcolor[HTML]{D55E00}{30.5} & \textcolor[HTML]{029E73}{34.4} & \textcolor[HTML]{D55E00}{31.5} & & 32.4 & \textbf{33.4} & 8.3 & 4.4 \\
  R@1/5\up{} & 35.5 & \textcolor[HTML]{029E73}{36.0} & \textcolor[HTML]{029E73}{36.2} & \textcolor[HTML]{029E73}{37.6} & \textcolor[HTML]{D55E00}{35.4} & \textcolor[HTML]{029E73}{36.1} & \textcolor[HTML]{D55E00}{35.5} & & 36.0 & \textbf{36.2} & 24.6 & 20.0 \\
  \bottomrule
  \multicolumn{13}{l}{\begin{minipage}{0.5\linewidth}
   \scriptsize \textcolor[HTML]{029E73}{green}/\textcolor[HTML]{D55E00}{orange}: step-wise improvement/regression.
  \end{minipage}} \\
 \end{tabular}
\end{table}

\subsection{Component Analysis}\label{sec:component}
We ablate the input modalities to support claim~(iv) that incorporating audio improves streaming localization and reduces redundancy.
We fine-tune each Phi-4-mm variant for 1 epoch on CMD-AD and evaluate on the CMD-AD val set.
The results are shown in \cref{tab:ablation-modality}.

\textbf{Modalities.}
We train three variants of the full vision-audio-text baseline, dropping video, audio, or both.
Removing video collapses CIDEr from 32.4 to 8.3, halves SODA, and pushes REP from 11.5\,\% to 49.6\,\%.
Removing audio improves the segment-level task but reduces streaming performance where streaming CIDEr drops from 10.7 to 9.7, OL worsens from 3.9\,\% to 5.3\,\%, and REP from 11.5\,\% to 16.3\,\%.
This suggests that audio is important for streaming and helps the model decide \emph{when} to speak rather than \emph{what} to say.
This confirms that video is the dominant signal and audio's contribution stacks on top of it, supporting claim~(iv).

\textbf{Augmentations.}
We apply input augmentations during training: 
\emph{segment} randomizes the pad around each clip; 
\emph{subtitle} overlays subtitle bars for robustness; 
\emph{face} drops character bounding boxes to match streaming evaluation; 
\emph{context} perturbs the AD context list by dropping or repeating entries; 
\emph{audio} applies random gain and Gaussian noise; and 
\emph{frame} combines aspect-ratio crops with photometric perturbations.
Dropping \emph{segment} jitter has the largest impact on segment-level CIDEr while reducing overlap and repetition.
Dropping \emph{audio} augmentation helps streaming and segment-level tasks.
Removing \emph{context} resampling sharply increases repetitions and degrades segment-level performance.
The other augmentations are inconclusive.
We remove \emph{segment} and \emph{audio} augmentations for training.

\section{Limitations}\label{sec:limitations}

Our work has the following limitations.
Currently, all models fall short of human-level AD quality and may produce errors that are confusing or misleading to visually impaired viewers.
Our benchmark, while broader in scope than prior datasets, does not yet cover the full diversity of video genres, languages, and cultural contexts found in the real world.
Because our benchmark is built on YouTube videos, content may become unavailable over time.
We mitigate this with our alignment process and regular availability scans but reproducibility cannot be fully guaranteed.
Our streaming evaluation uses the verified transcript, so ASR errors are not reflected in the reported numbers.
Finally, our metrics capture lexical and semantic overlap but not the perceived quality of AD for visually impaired audiences.
Alignment of these metrics with perceived quality on \audiodescbench{} remains to be verified through user studies with AD consumers and creators.

\section{Conclusion}
We investigated full-video AD generation for visually impaired audiences, a setting that prior work has mostly approached at the segment level with curated metadata and ground-truth timestamps.
We introduced \audiodescbench{}, a benchmark of long-form open-domain videos with manually verified transcripts and AD.
We reformulated AD generation as three tasks at increasing levels of difficulty: segment-level, streaming, and document-level.
We proposed two streaming baselines, the zero-shot \audiodesczeromodel{} and the fine-tuned \audiodescmodel{}, which jointly predict AD sentences and their timestamps from a sliding window over the video, transcript, and previously generated AD.
\audiodescmodel{} matches or exceeds prior methods on segment-level CMD-AD, MAD-Eval, and \audiodescbench{}, produces coherent AD across full-length videos without ground-truth boundaries, and runs faster than prior pipelines at \(<\)5B parameters.
Open challenges include entity recognition and grounding from the video alone, identifying character names, their belongings, and relations across minutes or hours, and suppressing repetitive or collapsed outputs without sacrificing streaming throughput.
Closing these gaps is what turns automated AD from a research artifact into a practical accessibility tool at the scale of online video.

\bibliographystyle{plainnat}
\bibliography{main}

\clearpage
\appendix

\section{\audiodescbench{} Dataset Construction}\label{sec:appendix-benchmark}

\begin{figure}[!t]
    \centering
    \begin{subfigure}[b]{0.45\textwidth}
        \centering
        \includegraphics[height=5cm,width=\linewidth,keepaspectratio]{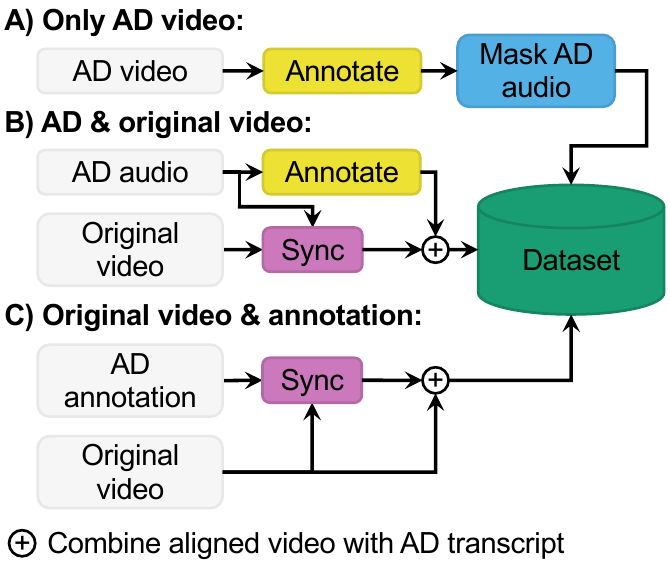}
        \caption{Alignment pipeline.}
        \label{fig:dataset-alignment}
    \end{subfigure}%
    \hfill
    \begin{subfigure}[b]{0.2\textwidth}
        \centering
        \includegraphics[height=4cm,width=\linewidth,keepaspectratio]{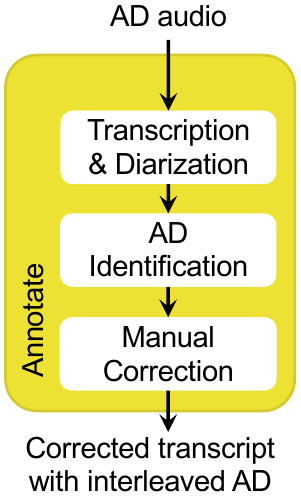}
        \caption{Annotation.}
        \label{fig:dataset-annotate}
    \end{subfigure}%
    \hfill 
    \begin{subfigure}[b]{0.35\textwidth}
        \centering
        \includegraphics[height=5cm,width=\linewidth,keepaspectratio]{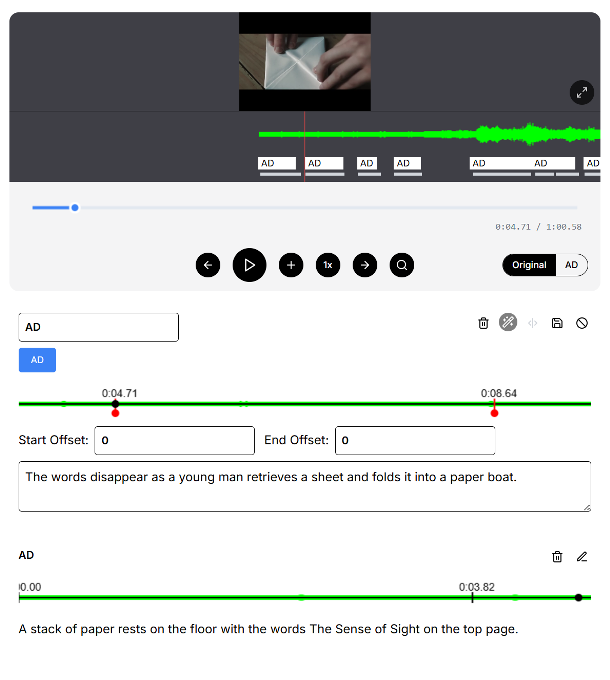}
        \caption{Annotation tool.}
        \label{fig:dataset-tool}
    \end{subfigure}
    \caption{\textbf{\audiodescbench{} construction pipeline.} (\subref{fig:dataset-alignment})~We source paired YouTube videos with original and AD-augmented audio tracks and align them via cross-correlation with \texttt{describealign}, comparing three alignment strategies. (\subref{fig:dataset-annotate})~We transcribe the AD track with Whisper~\citep{radford2023robust} and separate AD from speech using Silero VAD~\citep{silero} and pyannote diarization~\citep{plaquet2023pyannote}. (\subref{fig:dataset-tool})~We finalize timestamps and transcripts through manual review using a custom annotation tool.}
    \label{fig:dataset-creation}
\end{figure}

Constructing \audiodescbench{} requires pairing each full-length video with an aligned AD track.
We address this in three stages: video sourcing, AD extraction, and manual verification (\cref{fig:dataset-creation}).

\textbf{Video sourcing.}
We source videos from YouTube and prioritize pairs of the same content with two audio tracks, one original and one with AD, so that the raw audio is preserved for model input.
In case no paired videos exist, we resort to originals with a separate AD audio track or masked AD videos.

\textbf{Alignment, annotation, and verification.}
Following the annotation-process by \citet{han2024autoad3}, we first obtain candidate AD sentences and timestamps by transcribing the AD track with Whisper~\citep{radford2023robust}, leveraging Silero voice activity detection~\citep{silero} and using pyannote for speaker diarization~\citep{plaquet2023pyannote}.
We identify AD sentences by the lowest usage of personal pronouns~\citep{han2024autoad3}.
We align the original video to the annotated AD video via audio cross-correlation using \texttt{describealign}\footnote{\href{github.com/julbean/describealign}{https://github.com/julbean/describealign}}.
A custom tool then lets annotators correct temporal alignment and edit transcripts and AD sentences.
Each video is reviewed by one annotator and spot-checked by a second annotator for quality control.
Annotators corrected 73.8\,\% of all AD events and transcript segments.

\subsection{Reproducibility}

The benchmark is available as a list of YouTube IDs with timestamps and AD sentences.
To support reproducibility, our alignment process is even applicable if the original or AD video is no longer available.
We support three alignment strategies (\cref{fig:dataset-alignment}): audio-to-audio alignment (default) when both the original and AD videos are available, transcript alignment when the AD video is missing, where we transcribe the original audio and match it to the annotations, and masking when only the AD video is available, where AD segments are replaced with noise.
Note that masked samples should not be used for training to avoid bias.
The availability of each video will be checked by regularly scanning the benchmark.
For future comparability, we recommend publishing the predictions for each video.

\section{\audiodesczeromodel{} Prompts}\label{sec:appendix-zeroshot-prompts}

\audiodesczeromodel{} chains three prompted stages on each sliding window: (i) a video captioning stage, (ii) a decision stage, and (iii) an AD generation stage.
The face detection and character identification stage of AutoAD-Zero~\citep{xie2024autoadzero} is unchanged and prompt-free.
We list the prompt templates in \cref{fig:prompt-captioning,fig:prompt-decision,fig:prompt-ad}.

\begin{figure}[!t]
\centering
\begin{minipage}{0.95\linewidth}
\begin{lstlisting}
"""Video captioning from AutoAD-Zero"""

PROMPT_TEMPLATE = (
    "Please describe the movie clip in the following four steps: "
    "1. Identify main characters{char_text}; "
    "2. Describe the actions of characters in one sentence, i.e., who is doing what, focusing on the movements; "
    "3. Describe the interactions between characters in one sentence, such as looking; "
    "4. Describe the facial expressions of characters in one sentence. "
    "Note, colored boxes are provided for character indications only, DO NOT mention them in the description. "
    "Make sure you do not hallucinate information. "
    "Template: 1. Main characters: ''; 2. Actions: ''; 3. Character-character interactions: ''; 4. Facial expressions: ''."
)

# {char_text} is built from the face-detector output.
EXAMPLE_CHAR_TEXT_WITH_FACES = (
    " (labeled by boxes): Max Mustermann (red), Erika Mustermann (orange)"
)
EXAMPLE_CHAR_TEXT_NO_FACES = ""

RENDERED_EXAMPLE = PROMPT_TEMPLATE.format(char_text=EXAMPLE_CHAR_TEXT_WITH_FACES)
\end{lstlisting}
\end{minipage}
\caption{\textbf{Video captioning prompt.} This is the exact prompt used by AutoAD-Zero~\citep{xie2024autoadzero}.}
\label{fig:prompt-captioning}
\end{figure}

\begin{figure}[!t]
\centering
\begin{minipage}{0.95\linewidth}
\begin{lstlisting}
"""Decision prompt"""

PROMPT_TEMPLATE = (
    "You are deciding whether an audio description (AD) sentence should be inserted into a speech-free moment.\n\n"
    "{context_block}"
    "VISUAL DESCRIPTION (what is visible in the speech-free window):\n{visual_description}\n\n"
    "{gap_description}"
    "Should an audio description be inserted here? Answer YES only if the visuals contain meaningful "
    "new action, character movement, or narrative information that a blind viewer would benefit from "
    "and that is NOT already conveyed by RECENT AUDIO DESCRIPTIONS. "
    "Answer YES if the scene introduces a new action, location change, character behaviour, or important "
    "visual detail that the context does not yet describe (e.g. a character enters the room, picks something "
    "up, or the setting shifts).\n"
    "Answer NO if the scene is static or uneventful or if the context already explains what is happening "
    "(e.g. characters keep looking at each other, no new movement vs. the prior window).\n\n"
    "Answer with exactly one word: YES or NO."
)

# {gap_description}
GAP_DESCRIPTION_TEMPLATE = "A speech-free gap of {duration:.1f} seconds is available for narration.\n\n"
OPEN_GAP_DESCRIPTION = "An open-ended speech-free window is available for narration.\n\n"
EXAMPLE_GAP_DESCRIPTION = GAP_DESCRIPTION_TEMPLATE.format(duration=3.4)

# {context_block} is a multi-line block built from the recent transcript and recent AD predictions
CONTEXT_BLOCK_TEMPLATE = "RECENT DIALOGUE:\n{dialogue}\n\nRECENT AUDIO DESCRIPTIONS:\n{ads}\n\n"
EMPTY_DIALOGUE_PLACEHOLDER = "No recent dialogue."
EMPTY_ADS_PLACEHOLDER = "No recent audio descriptions."
EXAMPLE_CONTEXT_BLOCK = CONTEXT_BLOCK_TEMPLATE.format(
    dialogue="I just want to make a good impression.\nYou'll be fine, just be yourself.",
    ads="Max adjusts his collar nervously.\nErika squeezes his hand.",
)

# {visual_description}
EXAMPLE_VISUAL_DESCRIPTION = (
    "1. Main characters: Max Mustermann, Erika Mustermann; "
    "2. Actions: Max Mustermann is shaking Erika Mustermann's hand; "
    "3. Character-character interactions: Max Mustermann and Erika Mustermann look at each other; "
    "4. Facial expressions: Max Mustermann looks slightly nervous; Erika Mustermann appears cordial."
)

RENDERED_EXAMPLE = PROMPT_TEMPLATE.format(
    context_block=EXAMPLE_CONTEXT_BLOCK,
    visual_description=EXAMPLE_VISUAL_DESCRIPTION,
    gap_description=EXAMPLE_GAP_DESCRIPTION,
)
\end{lstlisting}
\end{minipage}
\caption{\textbf{Decision prompt.} Inserted before AD generation to predict whether the current window should produce an AD.}
\label{fig:prompt-decision}
\end{figure}

\begin{figure}[!t]
\centering
\begin{minipage}{0.95\linewidth}
\begin{lstlisting}
"""AD generation prompt based on AutoAD-Zero"""

# Verb priority list by AutoAD-Zero
VERB_LIST = "look, turn, take, hold, pull, walk, run, watch, stare, grab, fall, get, go, open, smile"

# Based on AutoAD-Zero with an additional context block
PROMPT_TEMPLATE = (
    "{context_block}"
    "Please summarise the following description for one movie clip into ONE succinct audio description (AD) sentence.\n"
    "Description: {description}\n\n"
    "Focus on the most attractive characters and their actions (focus on point 2., supplemented by point 3.).\n"
    "For characters, use their first names, remove titles such as 'Mr.' and 'Dr.'. "
    "If names are not available, use pronouns such as 'He' and 'her', do not use expression such as 'a man'.\n"
    "For actions, avoid mentioning the camera, and do not focus on 'talking' or position-related ones such as 'sitting' and 'standing'.\n"
    "Do not mention characters' mood.\n"
    "Do not hallucinate information that is not mentioned in the input.\n"
    "Do NOT restate or paraphrase any sentence already present in RECENT AUDIO DESCRIPTIONS; "
    "describe only what has changed or is new in the current clip.\n"
    "Try to identify the following motions (with decreasing priorities): {verb_list}, and use them in the description.\n"
    "Provide the AD from a narrator perspective and adjust the length of the output according to the duration.\n"
    "Duration of the video clip: {duration:.1f}s\n\n"
    "Template: AD: ''."
)

CONTEXT_BLOCK_TEMPLATE = "RECENT DIALOGUE:\n{dialogue}\n\nRECENT AUDIO DESCRIPTIONS:\n{ads}\n\n"
EXAMPLE_CONTEXT_BLOCK = CONTEXT_BLOCK_TEMPLATE.format(
    dialogue="I just want to make a good impression.",
    ads="Max adjusts his collar nervously.",
)

EXAMPLE_DESCRIPTION = (
    "1. Main characters: Max Mustermann, Erika Mustermann; "
    "2. Actions: Max Mustermann is shaking Erika Mustermann's hand; "
    "3. Character-character interactions: Max Mustermann and Erika Mustermann look at each other; "
    "4. Facial expressions: Max Mustermann looks slightly nervous; Erika Mustermann appears cordial."
)

# {duration} is the gap length in seconds, formatted with one decimal.
EXAMPLE_DURATION = 3.4

RENDERED_EXAMPLE = PROMPT_TEMPLATE.format(
    context_block=EXAMPLE_CONTEXT_BLOCK,
    description=EXAMPLE_DESCRIPTION,
    verb_list=VERB_LIST,
    duration=EXAMPLE_DURATION,
)
\end{lstlisting}
\end{minipage}
\caption{\textbf{AD generation prompt.} This is an extension for streaming of the the AD generation prompt of AutoAD-Zero~\citep{xie2024autoadzero}}
\label{fig:prompt-ad}
\end{figure}

\begin{table}[!t]
\centering
\caption{Full training, inference, and hardware configuration.}
\label{tab:training-details}
\small
\begin{tabular}{@{}lll@{}}
\toprule
\textbf{Category} & \textbf{Hyperparameter} & \textbf{Value} \\
\midrule
\multirow{2}{*}{Models}
  & Qwen-3.5    & \href{https://huggingface.co/Qwen/Qwen3.5-4B}{\texttt{Qwen/Qwen3.5-4B}} \\
  & Phi-4-mm    & \href{https://huggingface.co/microsoft/Phi-4-multimodal-instruct}{\texttt{microsoft/Phi-4-multimodal-instruct}} \\
\midrule
\multirow{2}{*}{Visual input}
  & Qwen-3.5 resolution      & up to \(720{\times}1280\) \\
  & Phi-4-mm resolution      & \(448{\times}448\) \\
\midrule
\multirow{11}{*}{Optimization}
  & Optimizer                & AdamW~\citep{loshchilov2017decoupled} \\
  & Loss                     & Cross-entropy \\
  & Weight decay             & 0.01 \\
  & Gradient clipping        & 1.0 \\
  & Epochs                   & \(\sim\)2 (with early stopping) \\
  & Total training steps     & 20{,}000 \\
  & Batch size               & 8 \\
  & LR schedule              & Cosine decay~\citep{loshchilov2016sgdr} \\
  & Warm-up steps            & 5\,\% of total steps \\
  & Random seed              & 42 \\
  & Early stopping           & Mean loss on CMD-AD val splits \\
\midrule
\multirow{4}{*}{Trainable components}
  & Vision projection layer  & Trained \\
  & Audio projection layer   & Trained \\
  & Language model           & LoRA adapters only \\
  & Encoders                 & Frozen \\
\midrule
\multirow{3}{*}{LoRA~\citep{hu2022lora}}
  & Rank \(r\)                 & 16 \\
  & Scaling \(\alpha\)         & 32 \\
  & Dropout                  & 0.0 \\
\midrule
\multirow{3}{*}{Learning rate}
  & Search method            & Optuna, 10 trials per backbone \\
  & Selection criterion      & Mean loss on CMD-AD val splits \\
  & LR                       & \(4.3\mathrm{e}{-5}\) (Qwen-3.5), \(9.2\mathrm{e}{-5}\) (Phi-4-mm)\\
\midrule
\multirow{3}{*}{Sample budget}
  & \texttt{WAIT} signals    & 25\,\% \\
  & Segment-level samples    & 37.5\,\% \\
  & Streaming samples        & 37.5\,\% \\
\midrule
\multirow{6}{*}{Segment-level task}
  & Decoding                 & Greedy \\
  & Max new tokens           & 128 \\
  & Padding \(\delta_p\)       & 2\,s \\
  & Eval padding \(\delta_p\)       & 2\,s (CMD-AD, \audiodescbench{}); 0\,s (MAD-Eval) \\
  & Conditioning             & Character names, face detections, \\
  &                          & ground-truth timestamps \\
\midrule
\multirow{8}{*}{Streaming task}
  & Decoding                 & Greedy \\
  & Max new tokens           & 128 \\
  & Presence penalty~\citep{keskar2019ctrl} &1.5 \\
  & Advance step \(\delta\)     & 4\,s \\
  & Visual window size \(w_v\)  & 8\,s \\
  & Text context window \(w_c\) & 120\,s \\
  & Conditioning                & Previous ADs, ground-truth transcript \\
\midrule
\multirow{7}{*}{Hardware}
  & Training nodes           & 1 \\
  & Training GPUs            & 8\(\times\) NVIDIA A100 \\
  & Training CPU             & AMD Epyc \\
  & Inference GPU            & 1\(\times\) NVIDIA A100 \\
  & Inference CPU            & Intel Xeon CPU \\
  & Training time           & \(\sim\)160\,GPU-h (Qwen-3.5),\\
  &                          & \(\sim\)80\,GPU-h (Phi-4-mm) \\
\bottomrule
\end{tabular}
\end{table}

\section{Detailed Experimental Setup}\label{sec:appendix-detailed-setup}

We provide hyperparameters and other details required for reproduction of \audiodescmodel{} in the \cref{tab:training-details}.

\subsection{CMD-AD Validation Split}\label{sec:cmd-ad-val-split}

CMD-AD only has a train and test split, but we need a validation split for model development and hyperparameter tuning.
We create a validation split by randomly sampling 25 movies from the training set and select the corresponding clips and segments.
The selected movies are: \texttt{tt0020629}, \texttt{tt0054331}, \texttt{tt0073802}, \texttt{tt0088846}, \texttt{tt0096101}, \texttt{tt0101635}, \texttt{tt0108065}, \texttt{tt0116209}, \texttt{tt0120631}, \texttt{tt0151568}, \texttt{tt0232500}, \texttt{tt0285742}, \texttt{tt0332452}, \texttt{tt0376541}, \texttt{tt0421729}, \texttt{tt0463985}, \texttt{tt0800039}, \texttt{tt0959337}, \texttt{tt1186830}, \texttt{tt1355644}, \texttt{tt1622979}, \texttt{tt1981677}, \texttt{tt2493486}, \texttt{tt3531824}, and \texttt{tt5113040}.

\subsection{Learning Rate Sweep}\label{sec:optuna-lr}

\begin{figure}[!t]
    \centering
    \begin{subfigure}[t]{\linewidth}
        \centering
        \includegraphics[width=\linewidth]{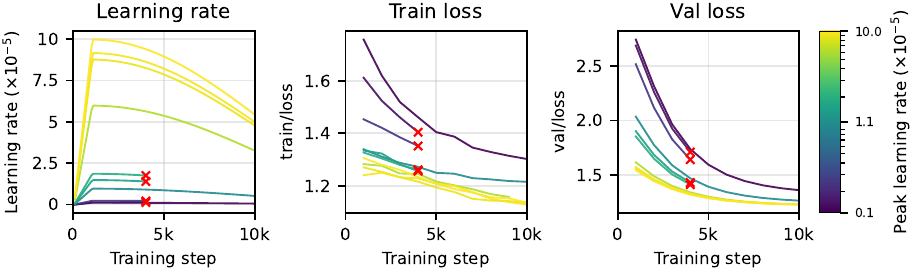}
        \caption{Phi-4-mm}
        \label{fig:lr-sweep-phi4}
    \end{subfigure}
    \par\medskip
    \begin{subfigure}[t]{\linewidth}
        \centering
        \includegraphics[width=\linewidth]{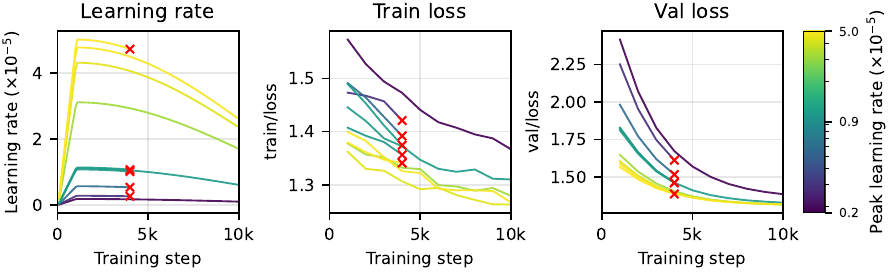}
        \caption{Qwen-3.5}
        \label{fig:lr-sweep-qwen}
    \end{subfigure}
    \caption{\textbf{Learning rate sweep.} CMD-AD validation loss across 10 Optuna trials per model.}
    \label{fig:lr-sweep}
\end{figure}

We use Optuna~\citep{akiba2019optuna} to search for the best learning rate for both Phi-4-mm and Qwen-3.5 with 10 trials each, evaluating the validation loss on our CMD-AD val split.
The search space samples the learning rate log-uniformly from \(1\mathrm{e}{-6}\) to \(1\mathrm{e}{-4}\).
\cref{fig:lr-sweep} shows the trial results for both models.
The chosen learning rates are \(9.2\mathrm{e}{-5}\) for Phi-4-mm and \(4.3\mathrm{e}{-5}\) for Qwen-3.5.

\begin{table}[!t]
 \centering
 \caption{Effect of evaluating published predictions against missing entries on CMD-AD.}
 \label{tab:appendix-dropmissing}
 \footnotesize
 \setlength{\tabcolsep}{2pt}
 \begin{tabular}{lccccc}
  \toprule
  & \multicolumn{5}{c}{CMD-AD} \\
  \cmidrule(lr){2-6}
  Method & CIDEr\up{} & R@1/5\up{} & CRITIC\up{} & ActionScore\up{} & LLM-AD-Eval\(^{\ddagger}\)\up{} \\
  \midrule
  AutoAD-III~\citep{han2024autoad3} (as reported in the paper) & 25.0 & 31.2 & 32.7 & 31.5 & 2.89 | 2.01 \\
  AutoAD-III~\citep{han2024autoad3} (published preds, full) & 25.1 & 31.1 & 31.8 & 29.7 & 3.02 | 2.16 \\
  AutoAD-III~\citep{han2024autoad3} (published preds, drop missing) & 24.9 & 31.2 & 32.4 & 30.0 & 3.03 | 2.18 \\
  AutoAD-III~\citep{han2024autoad3}\(^{\dagger}\) & 21.6 & 30.2 & 26.7 & 32.9 & 3.26 | 2.35 \\
  \midrule
  AutoAD-Zero~\citep{xie2024autoadzero} (as reported in the paper) & 17.7 & 26.9 & 43.7 & 25.5 & 2.83 | 1.96 \\
  AutoAD-Zero~\citep{xie2024autoadzero} (published preds, full) & 17.7 & 27.5 & 38.5 & 25.5 & 2.91 | 2.08 \\
  AutoAD-Zero~\citep{xie2024autoadzero} (published preds, drop missing) & 17.7 & 27.6 & 38.0 & 25.7 & 2.92 | 2.09 \\
  \midrule
  Shot-by-shot\(^{\star}\)~\citep{xie2025shotbyshot} (as reported in the paper) & 26.1 & 36.5 & 49.1 & 32.5 & 3.17 | 2.66 \\
  Shot-by-shot\(^{\star}\)~\citep{xie2025shotbyshot} (published preds, full) & 26.1 & 34.8 & 43.2 & 32.5 & 3.34 | 2.80 \\
  Shot-by-shot\(^{\star}\)~\citep{xie2025shotbyshot} (published preds, drop missing) & 25.8 & 34.9 & 42.5 & 32.7 & 3.34 | 2.81 \\
  \midrule
  Shot-by-shot~\citep{xie2025shotbyshot} (as reported in the paper) & 26.3 & 33.0 & 47.8 & 28.4 & 3.15 | 2.42 \\
  Shot-by-shot~\citep{xie2025shotbyshot} (published preds, full) & 26.3 & 31.4 & 43.3 & 28.4 & 3.33 | 2.55 \\
  Shot-by-shot~\citep{xie2025shotbyshot} (published preds, drop missing) & 25.9 & 31.4 & 42.9 & 28.5 & 3.33 | 2.56 \\
  \bottomrule
 \end{tabular}
\end{table}

\section{Metrics}\label{sec:appendix-metrics}

\subsection{Segment-level metrics}

\textbf{CIDEr}~\citep{vedantam2015cider} measures n-gram overlap with the ground-truth AD while down-weighting n-grams that are common across the full ground-truth corpus.
Scores are computed for n-grams of length 1 to 4 and averaged.

\textbf{Recall@\(k\)/\(N\)}~\citep{han2023autoad2} addresses the imperfect semantic alignment between predicted and ground-truth AD by scoring each predicted sentence against a window of \(N\) neighboring ground-truth sentences using BERTScore~\citep{zhang2019bertscore} and reporting the fraction of predictions that match within the top-\(k\).
The metric is not fully specified with respect to averaging strategy and ordering, which leads to discrepancies across implementations (see \cref{sec:appendix-metric-reproduction}).

\textbf{CRITIC}~\citep{han2024autoad3} calculates the IoU between character identities in predictions and references extracted with a co-referencing model and the ground-truth character list per movie.

\textbf{LLM-AD-Eval}~\citep{han2024autoad3} provides a complementary assessment that is less sensitive to exact wording using an LLM as a judge, since AD datasets lack diverse reference captions.
The LLM rates how well the prediction matches the reference on a scale of 1 to 5.
The final score is the micro-average across all prediction-reference pairs.
We use the prompts by \citet{xie2024autoadzero} and the language models \texttt{meta-llama/Llama-2-7b-chat-hf}~\citep{touvron2023llama} and \texttt{meta-llama/Meta-Llama-3-8B-Instruct}~\citep{grattafiori2024llama}.
We use vLLM~\citep{kwon2023efficient} to run the LLMs.

\subsection{Streaming metrics}

For streaming evaluation we group metrics into four categories: quality, temporal localization, diagnostics, and efficiency.

\subsubsection{Quality}\label{sec:appendix-metrics-quality}

Quality metrics measure how well each AD is phrased, for which we adopt the standard dense video captioning metrics~\citep{krishna2017dense, yang2023vid2seq, zhou2024streaming}.
All metrics are averaged over tIoU thresholds 0.1, 0.3, and 0.5, dropping the 0.9 used in dense video captioning since AD timing is loose by design.

\textbf{SODA}~\citep{fujita2020soda} computes an F-measure over a temporal matching, using METEOR~\citep{banerjee2005meteor} as the caption similarity score.
The temporal matching is found with dynamic programming and a tIoU-based cost matrix, where a tIoU matching threshold is applied to zero non-matching entries.
SODA only reaches a high score when AD events are both well described and well localized.

\textbf{CIDEr}~\citep{vedantam2015cider} is calculated in the same way as for the segment-level task but calculated and averaged for each tIoU threshold.

\textbf{METEOR}~\citep{banerjee2005meteor} scores a caption by aligning its unigrams to the reference using exact, stem, synonym, and paraphrase matches, then computing a F-measure and a penalty that rewards contiguous matched blocks.

\textbf{ActionScore}~\citep{xie2025shotbyshot} computes the recall of the actions extracted by a LLM from the ground-truth ADs. 
A text embedding model is used to calculate the cosine similarity of each action to the prediction to identify the actions.

\subsubsection{Temporal localization}

Temporal localization metrics capture whether ADs are emitted at the right time, for which we report Recall and Precision over the tIoU-matched pairs used for quality metrics (see \cref{sec:appendix-metrics-quality}).

\textbf{Recall} measures the fraction of ground-truth ADs that are matched by a predicted AD under the tIoU thresholds.

\textbf{Precision} measures the fraction of predicted ADs that are matched to a ground-truth AD under the tIoU thresholds.

\subsubsection{Diagnostics}

Diagnostic metrics target streaming-specific failure modes, for which we report REP and OL.

\textbf{REP} or repetition is the percentage of time-ordered consecutive AD pairs \((p_i, p_{i+1})\) across all videos for which \(\text{METEOR}(p_i, p_{i+1}) = 1\), i.e., exact duplicates.

\textbf{OL} or overlap is the fraction of the total predicted AD duration overlapping transcribed speech.

\subsubsection{Efficiency}

Efficiency metrics reflect runtime cost, for which we report the real time factor (RTF) and lag over all AD-emitting windows.
Thus, \texttt{WAIT} signals or skipping transcript is not measured.
Efficiency measurements must be run on a single A100 GPU with a batch size of one.

\textbf{RTF} is the average ratio \(\text{RTF} = \frac{1}{N} \sum_{i=1}^{N} t^{\text{wall}}_{i} / \delta_{i}\) between per-window generation latency \(t^{\text{wall}}_{i}\) and the streaming advance step \(\delta_{i}\), with \(\text{RTF} < 1\) meaning the model runs faster than real time.
Note that a window only counts to \(N\) if it predicts an actual AD event.

\textbf{LAG} is the average \(\text{LAG} = \frac{1}{N} \sum_{i=1}^{N} (t_i - t^{\text{start}}_i)\), where \(t^{\text{start}}_i\) is the start timestamp of the first newly emitted AD in window \(i\) and \(t_i\) is its emission time or cursor position.

\subsection{Reproducibility of Metrics}\label{sec:appendix-metric-reproduction}

Some metrics are shared as loose code snippets, which compromises reproducibility.
We validate our metric implementation by running published predictions against the published results.
Results for AutoAD-III~\citep{han2024autoad3}, AutoAD-Zero~\citep{xie2024autoadzero} and Shot-by-shot~\citep{xie2025shotbyshot} are available for CMD-AD.
Some published results for AutoAD-III and AutoAD-Zero are reported by \citet{xie2025shotbyshot}.
The results are shown in \cref{tab:appendix-dropmissing}.
We see that Recall@1/5~\citep{han2023autoad2}, CRITIC~\citep{han2024autoad3}, and LLM-AD-Eval~\citep{han2024autoad3} differ across models.

We found that Recall@1/5 predictions must be sorted first by IMDb ID (movie), then by YouTube ID, and finally by start timestamp, yielding a single, correctly-ordered sequence of predictions and references across all videos and clips.
Clips that have fewer ground-truth AD events than \(N\) are skipped.
The micro-average over all windows is reported as the final score.
This can yield unwanted boundary effects.
Other approaches may use a different ordering or python package versions affecting the final score.

For CRITIC, we achieve a reasonably close score for AutoAD-III~\citep{han2024autoad3}, the work that introduced the metric.
The gap is considerably larger for Shot-by-shot~\citep{xie2025shotbyshot}, where we obtain 43.2 against the reported 49.1.
Since the CIDEr scores match, issues with the published predictions are unlikely.
More plausible causes are again differences in averaging or python package versions.

Our LLM-AD-Eval implementation tends to achieve slightly higher scores than those reported for Llama-2-7B and Llama-3-8B.
The reason is likely differences in language model execution such as numerical precision, hardware, or frameworks.

In summary, our implementation reproduces published results closely enough to validate its correctness, while the residual discrepancies illustrate why loose code snippets are insufficient for reliable benchmarking.
This motivates the unified, executable evaluation suite we release with \audiodescbench{}.

\section{Impact of an Incomplete CMD-AD Test Set}~\label{sec:appendix-incomplete-cmdad}

The CMD-AD test split is based on 98 movies and originally contains 591 clips with 7{,}316 AD events~\citep{bain2020CMD, han2024autoad3}.
We evaluate on the 551 clips (93.2\,\%, 6{,}740 AD events) still available at the time of publication.
To validate the impact of the incomplete test set, we compare complete against incomplete predictions using the setup described in \cref{sec:appendix-metric-reproduction}.
The results are reported in \cref{tab:appendix-dropmissing}.
The CIDEr score drops by -0.2 for AutoAD-III~\citep{han2024autoad3}, 0.0 for AutoAD-Zero~\citep{xie2024autoadzero}, and -0.4 for Shot-by-shot~\citep{xie2025shotbyshot}.
CRITIC is affected by +0.6 for AutoAD-III, -0.5 for AutoAD-Zero, and -0.4 for Shot-by-shot.
Recall@1/5, ActionScore and LLM-AD-Eval are only minimally or not affected.
Across all metrics, the differences between complete and incomplete evaluation are small and do not change the relative ranking of methods, confirming that results obtained on the available 551 clips remain comparable to those reported on the full CMD-AD test split.

\section{Extended Segment-level Results}\label{sec:appendix-extended-segment-results}

We provide extended results for the segment-level tasks on CMD-AD~\citep{bain2020CMD, han2024autoad3}, MAD-Eval~\citep{soldan2022MAD,han2023autoad} and \audiodescbench{} in \cref{tab:segment_cmdad,tab:segment_madeval,tab:segment_audiodescbench}.
Specifically, the tables report the ActionScore for CMD-AD, MAD-Eval, and LLM-AD-Eval for each dataset.
\audiodescmodel{} models with the \textit{randpad} tag are trained with random padding \(\delta_p \sim \mathcal{U}(0, 2)\,\text{s}\) for the segment-level task instead of fixed padding \(\delta_p = 2\,\text{s}\).

\begin{table}[!t]
 \centering
 \caption{
  Segment-level evaluation on CMD-AD.
 }
 \label{tab:segment_cmdad}
 \footnotesize
 \setlength{\tabcolsep}{2pt}
 \begin{tabular}{clccccc}
  \toprule
  & & \multicolumn{5}{c}{CMD-AD} \\
  \cmidrule(lr){3-7}
  & Method & CIDEr\up{} & R@1/5\up{} & CRITIC\up{} & ActionScore\up{} & LLM-AD-Eval\(^{\ddagger}\)\up{} \\
  \midrule
  \multirow{6}{*}{\rotatebox[origin=c]{90}{Fine-tuned}} & AutoAD-II~\citep{han2023autoad2} & 13.5 & 26.1 & 8.2 &  & 2.08 | \phantom{0.00} \\
   & AutoAD-III~\citep{han2024autoad3} & 25.0 & 31.2 & 32.7 & 31.5 & 2.89 | 2.01 \\
   & DistinctAD~\citep{fang2025distinctad} & 22.7 & 33.0 &  &  & 2.88 | 2.03 \\
  \cmidrule(lr){2-7}
   & AutoAD-III~\citep{han2024autoad3}\(^{\dagger}\) & 21.6 & 30.2 & 26.7 & 32.9 & 3.26 | 2.35 \\
   & Phi-4-mm (ours) & 29.8 & 35.6 & 29.6 & 34.7 & 3.35 | 2.51 \\
   & Qwen-3.5 (ours) & \textbf{36.3} & \textbf{38.0} & 31.8 & \textbf{37.7} & 3.44 | 2.74 \\
  \midrule
  \multirow{4}{*}{\rotatebox[origin=c]{90}{Zero-shot}} & AutoAD-Zero~\citep{xie2024autoadzero} & 17.7 & 26.9 & 43.7 & 25.5 & 2.83 | 1.96 \\
   & Shot-by-shot~\citep{xie2025shotbyshot} & 26.3 & 33.0 & 47.8 & 28.4 & 3.15 | 2.42 \\
   & Shot-by-shot\(^{\star}\)~\citep{xie2025shotbyshot} & 26.1 & 36.5 & \textbf{49.1} & 32.5 & 3.17 | 2.66 \\
  \cmidrule(lr){2-7}
   & Qwen-3.5 (ours) & 18.8 & 29.8 & 44.0 & 32.0 & \textbf{3.48} | \textbf{2.86} \\
  \bottomrule
 \end{tabular}
\end{table}

\begin{table}[!t]
 \centering
 \caption{
  Segment-level evaluation on MAD-Eval.
 }
 \label{tab:segment_madeval}
 \footnotesize
 \setlength{\tabcolsep}{2pt}
 \begin{tabular}{clccccc}
  \toprule
  & & \multicolumn{4}{c}{MAD-Eval} \\
  \cmidrule(lr){3-6}
  & Method & CIDEr\up{} & R@5/16\up{} & ActionScore\up{} & LLM-AD-Eval\(^{\ddagger}\)\up{} \\
  \midrule
  \multirow{8}{*}{\rotatebox[origin=c]{90}{Fine-tuned}} & AutoAD-II~\citep{han2023autoad2} & 19.5 & 51.3 &  &  \\
   & AutoAD-III~\citep{han2024autoad3} & 24.0 & 52.8 &  &  \\
   & MovieSeq~\citep{lin2024learning} & 24.4 & 51.6 &  &  \\
   & DistinctAD~\citep{fang2025distinctad} & 27.3 & 56.0 &  &  \\
   & UniAD~\citep{wang2025uniad} & \textbf{28.2} & 54.9 &  &  \\
  \cmidrule(lr){2-6}
   & AutoAD-III~\citep{han2024autoad3}\(^{\dagger}\) & 21.7 & 51.1 & 33.8 & 3.30 | 2.44 \\
   & \audiodescmodel{} (Phi-4-mm, ours) & 19.4 & 49.9 & 32.5 & 2.89 | 2.24 \\
   & \audiodescmodel{} (Qwen-3.5, ours) & 24.9 & 54.8 & \textbf{39.3} & 3.08 | 2.60 \\
  \midrule
  \multirow{8}{*}{\rotatebox[origin=c]{90}{Zero-shot}} & MM-Narrator\(^{\star}\)~\citep{zhang2024mmnarrator} & 13.9 & 49.0 &  &  \\
   & LLM-AD\(^{\star}\)~\citep{chu2024llm} & 20.5 &  &  &  \\
   & AutoAD-Zero~\citep{xie2024autoadzero} & 22.4 & 47.0 &  &  \\
   & Shot-by-shot~\citep{xie2025shotbyshot} & 25.0 & 50.6 &  &  \\
   & StoryContext-AD~\citep{yang2025story} & 26.0 & 50.3 &  &  \\
   & NarrAD\(^{\star}\)~\citep{park2025narrad} & 26.4 & 54.0 &  &  \\
   & Shot-by-shot\(^{\star}\)~\citep{xie2025shotbyshot} & 26.9 & 56.4 &  &  \\
  \cmidrule(lr){2-6}
   & \audiodesczeromodel{} (Qwen-3.5, ours) & 22.0 & \textbf{57.1} & 34.6 & \textbf{3.52} | \textbf{3.12} \\
  \bottomrule
 \end{tabular}
\end{table}

\begin{table}[!t]
 \centering
 \caption{
  Segment-level evaluation on \audiodescbench{}.
 }
 \label{tab:segment_audiodescbench}
 \footnotesize
 \setlength{\tabcolsep}{2pt}
 \begin{tabular}{lccc}
  \toprule
  & \multicolumn{3}{c}{\audiodescbench{}} \\
  \cmidrule(lr){2-4}
  Method & CIDEr\up{} & R@5/16\up{} & LLM-AD-Eval\(^{\ddagger}\)\up{} \\
  \midrule
  AutoAD-III~\citep{han2024autoad3}\(^{\dagger}\) & 17.5 & 45.2 & 2.96 | 2.07 \\
   \audiodescmodel{} (Phi-4-mm, ours) & 32.3 & 52.6 & 3.13 | 2.33 \\
   \audiodescmodel{} (Qwen-3.5, ours) & \textbf{51.0} & \textbf{57.9} & 3.26 | 2.66 \\
  \midrule
   \audiodesczeromodel{} (Qwen-3.5, ours) & 12.6 & 45.8 & \textbf{3.41} | \textbf{2.80} \\
  \bottomrule
 \end{tabular}
\end{table}

\section{Per-Genre Results for \audiodescbench{}}\label{sec:appendix-genre-results}

To provide insights into per-genre performance on \audiodescbench{}, we report evaluation results for each genre subset for the segment-level task in \cref{tab:segment_per_genre} and for the streaming task in \cref{tab:streaming_per_genre}.

\begin{table}[!t]
 \centering
 \caption{Segment-level evaluation on \audiodescbench{} broken down by genre.}
 \label{tab:segment_per_genre}
 \footnotesize
 \setlength{\tabcolsep}{2pt}
 \begin{tabular}{lccc}
  \toprule
  & \multicolumn{3}{c}{\audiodescbench{}} \\
  \cmidrule(lr){2-4}
  Method & CIDEr\up{} & R@5/16\up{} & LLM-AD-Eval\(^{\ddagger}\)\up{} \\
  \midrule
  \multicolumn{4}{l}{\emph{Documentary (8 videos, 979 GT events)}} \\
  \midrule
  \audiodesczeromodel{} (Qwen-3.5, zero-shot) & 8.9 & 43.3 & \textbf{3.38} | \textbf{2.77} \\
  AutoAD-III~\citep{han2024autoad3} & 17.0 & 45.5 & 2.87 | 1.89 \\
  \audiodescmodel{} (Phi-4-mm, randpad) & 27.2 & 50.2 & 3.25 | 2.41 \\
  \audiodescmodel{} (Phi-4-mm) & 39.4 & 53.2 & 2.98 | 2.12 \\
  \audiodescmodel{} (Qwen-3.5, randpad) & 57.2 & 55.4 & 3.25 | 2.47 \\
  \audiodescmodel{} (Qwen-3.5) & \textbf{65.0} & \textbf{57.2} & 3.09 | 2.36 \\
  \midrule
  \multicolumn{4}{l}{\emph{Game (8 videos, 2281 GT events)}} \\
  \midrule
  \audiodesczeromodel{} (Qwen-3.5, zero-shot) & 11.7 & 44.9 & 3.37 | \textbf{2.65} \\
  AutoAD-III~\citep{han2024autoad3} & 15.8 & 44.2 & 3.04 | 2.04 \\
  \audiodescmodel{} (Phi-4-mm, randpad) & 19.6 & 48.9 & 3.36 | 2.37 \\
  \audiodescmodel{} (Phi-4-mm) & 25.6 & 48.9 & 3.13 | 2.19 \\
  \audiodescmodel{} (Qwen-3.5, randpad) & 33.4 & 54.0 & \textbf{3.40} | 2.63 \\
  \audiodescmodel{} (Qwen-3.5) & \textbf{40.3} & \textbf{56.4} & 3.27 | 2.52 \\
  \midrule
  \multicolumn{4}{l}{\emph{Movie (5 videos, 4159 GT events)}} \\
  \midrule
  \audiodesczeromodel{} (Qwen-3.5, zero-shot) & 12.6 & 46.6 & \textbf{3.46} | \textbf{2.83} \\
  AutoAD-III~\citep{han2024autoad3} & 18.8 & 45.3 & 2.96 | 2.12 \\
  \audiodescmodel{} (Phi-4-mm, randpad) & 21.3 & 48.5 & 3.30 | 2.49 \\
  \audiodescmodel{} (Phi-4-mm) & 31.7 & 52.0 & 3.04 | 2.14 \\
  \audiodescmodel{} (Qwen-3.5, randpad) & 45.8 & 56.6 & 3.36 | 2.81 \\
  \audiodescmodel{} (Qwen-3.5) & \textbf{52.4} & \textbf{58.2} & 3.25 | 2.70 \\
  \midrule
  \multicolumn{4}{l}{\emph{Performance (1 video, 596 GT events)}} \\
  \midrule
  AutoAD-III~\citep{han2024autoad3} & 12.4 & 41.6 & 2.98 | 2.20 \\
  \audiodesczeromodel{} (Qwen-3.5, zero-shot) & 13.3 & 44.5 & \textbf{3.80} | 3.50 \\
  \audiodescmodel{} (Phi-4-mm, randpad) & 30.1 & 49.6 & 3.70 | 3.39 \\
  \audiodescmodel{} (Phi-4-mm) & 37.8 & 51.8 & 3.62 | 3.21 \\
  \audiodescmodel{} (Qwen-3.5, randpad) & 47.7 & 56.1 & 3.76 | \textbf{3.57} \\
  \audiodescmodel{} (Qwen-3.5) & \textbf{56.2} & \textbf{57.0} & 3.69 | 3.49 \\
  \midrule
  \multicolumn{4}{l}{\emph{Short Film (11 videos, 1116 GT events)}} \\
  \midrule
  \audiodesczeromodel{} (Qwen-3.5, zero-shot) & 14.1 & 47.4 & 3.17 | 2.64 \\
  AutoAD-III~\citep{han2024autoad3} & 18.0 & 47.9 & 2.86 | 2.02 \\
  \audiodescmodel{} (Phi-4-mm, randpad) & 22.6 & 55.3 & \textbf{3.35} | 2.64 \\
  \audiodescmodel{} (Phi-4-mm) & 29.7 & 55.3 & 3.13 | 2.39 \\
  \audiodescmodel{} (Qwen-3.5, randpad) & 40.6 & 58.6 & \textbf{3.35} | \textbf{2.71} \\
  \audiodescmodel{} (Qwen-3.5) & \textbf{49.3} & \textbf{60.5} & 3.23 | 2.61 \\
  \bottomrule
 \end{tabular}
\end{table}

\begin{table}[!t]
 \centering
 \caption{Streaming evaluation on \audiodescbench{} broken down by genre.}
 \label{tab:streaming_per_genre}
 \footnotesize
 \setlength{\tabcolsep}{3pt}
 \begin{tabular}{lccccccc}
  \toprule
  & \multicolumn{3}{c}{Quality} & \multicolumn{2}{c}{Localization} & \multicolumn{2}{c}{Diagnostics} \\
  \cmidrule(lr){2-4} \cmidrule(lr){5-6} \cmidrule(lr){7-8}
  Method & SODA\up{} & CIDEr\up{} & METEOR\up{} & R\up{} & P\up{} & REP\down{} & OL\down{} \\
  \midrule
  \multicolumn{8}{l}{\emph{Documentary (8 videos, 979 GT events)}} \\
  \midrule
  \audiodesczeromodel{} (Qwen-3.5, zero-shot) & 0.6 & 7.3 & 4.6 & 18.5 & 44.3 & 2.3 & 30.3 \\
  \audiodescmodel{} (Phi-4-mm) & 1.6 & 28.8 & 6.9 & 41.7 & 52.1 & 15.7 & 3.3 \\
  \audiodescmodel{} (Phi-4-mm, randpad) & 1.7 & 23.5 & 5.7 & 44.1 & 50.5 & 18.7 & \textbf{2.3} \\
  \audiodescmodel{} (Qwen-3.5, randpad) & 2.3 & 37.8 & 7.6 & 45.3 & \textbf{55.3} & 0.2 & 2.7 \\
  \audiodescmodel{} (Qwen-3.5) & \textbf{2.6} & \textbf{54.4} & \textbf{10.0} & \textbf{47.5} & 54.5 & \textbf{0.0} & 2.4 \\
  \midrule
  \multicolumn{8}{l}{\emph{Game (8 videos, 2281 GT events)}} \\
  \midrule
  \audiodesczeromodel{} (Qwen-3.5, zero-shot) & 0.7 & 8.1 & 4.3 & 24.3 & 46.4 & 2.4 & 12.2 \\
  \audiodescmodel{} (Phi-4-mm) & 1.5 & 14.6 & 4.8 & 50.0 & 56.6 & 12.7 & 2.3 \\
  \audiodescmodel{} (Phi-4-mm, randpad) & 2.0 & 21.9 & 6.6 & 50.0 & 55.9 & 16.3 & \textbf{1.8} \\
  \audiodescmodel{} (Qwen-3.5, randpad) & 2.1 & 22.6 & \textbf{8.2} & 51.6 & 60.6 & 0.4 & 3.3 \\
  \audiodescmodel{} (Qwen-3.5) & \textbf{2.3} & \textbf{40.2} & 7.7 & \textbf{53.2} & \textbf{61.0} & \textbf{0.2} & 2.3 \\
  \midrule
  \multicolumn{8}{l}{\emph{Movie (5 videos, 4159 GT events)}} \\
  \midrule
  \audiodesczeromodel{} (Qwen-3.5, zero-shot) & 1.2 & 8.4 & 5.0 & 40.3 & 56.5 & 2.7 & 5.2 \\
  \audiodescmodel{} (Phi-4-mm, randpad) & 1.6 & 8.7 & 4.6 & \textbf{66.4} & 59.1 & 27.8 & 1.9 \\
  \audiodescmodel{} (Phi-4-mm) & 1.7 & 10.7 & 4.7 & 65.3 & 59.0 & 22.2 & 2.1 \\
  \audiodescmodel{} (Qwen-3.5, randpad) & \textbf{2.2} & 17.8 & \textbf{6.5} & 64.6 & \textbf{60.9} & \textbf{0.3} & 1.1 \\
  \audiodescmodel{} (Qwen-3.5) & \textbf{2.2} & \textbf{18.4} & 6.2 & 65.4 & 60.7 & 0.8 & \textbf{0.9} \\
  \midrule
  \multicolumn{8}{l}{\emph{Performance (1 video, 596 GT events)}} \\
  \midrule
  \audiodesczeromodel{} (Qwen-3.5, zero-shot) & 1.6 & 7.5 & \textbf{5.8} & 50.8 & \textbf{44.4} & 13.7 & 3.3 \\
  \audiodescmodel{} (Phi-4-mm, randpad) & 1.6 & 6.3 & 4.6 & 58.1 & 39.9 & 39.4 & 1.4 \\
  \audiodescmodel{} (Phi-4-mm) & 1.8 & 6.0 & 4.8 & 56.2 & 39.1 & 43.6 & 1.5 \\
  \audiodescmodel{} (Qwen-3.5) & 2.0 & 10.0 & 5.3 & 58.1 & 39.5 & \textbf{3.0} & \textbf{1.3} \\
  \audiodescmodel{} (Qwen-3.5, randpad) & \textbf{2.1} & \textbf{10.1} & 5.4 & \textbf{59.6} & 40.5 & 3.3 & 1.9 \\
  \midrule
  \multicolumn{8}{l}{\emph{Short Film (11 videos, 1116 GT events)}} \\
  \midrule
  \audiodesczeromodel{} (Qwen-3.5, zero-shot) & 1.6 & 12.7 & 5.7 & 41.5 & \textbf{62.5} & 2.5 & 4.5 \\
  \audiodescmodel{} (Phi-4-mm) & 1.7 & 11.7 & 4.7 & 70.4 & 58.2 & 13.4 & 1.1 \\
  \audiodescmodel{} (Qwen-3.5, randpad) & 2.1 & 17.8 & 5.6 & 67.3 & 58.4 & \textbf{0.0} & 0.7 \\
  \audiodescmodel{} (Phi-4-mm, randpad) & 2.2 & 17.4 & 5.8 & \textbf{71.3} & 58.5 & 19.7 & 0.8 \\
  \audiodescmodel{} (Qwen-3.5) & \textbf{2.6} & \textbf{23.9} & \textbf{6.3} & 70.4 & 58.0 & 0.3 & \textbf{0.5} \\
  \bottomrule
 \end{tabular}
\end{table}

\section{Qualitative Examples}\label{sec:appendix-qualitative}

\begin{figure}[!t]
    \centering
    \begin{subfigure}[t]{\linewidth}
        \centering
        \includegraphics[width=\linewidth]{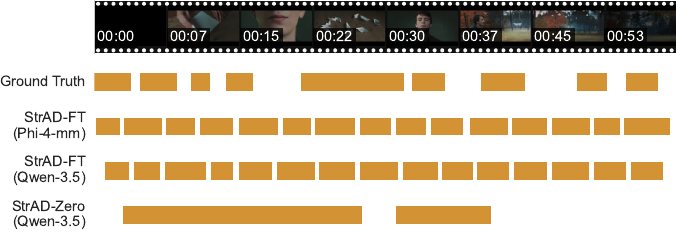}
        \caption{\emph{The Sense of Sight}. Original video: \href{https://youtu.be/Oph8kL_Z4c4}{\nolinkurl{youtu.be/Oph8kL_Z4c4}}; audio-described version: \href{https://youtu.be/fNV8TUqjfbQ}{\nolinkurl{youtu.be/fNV8TUqjfbQ}}.}
        \label{fig:supp-timeline-a}
    \end{subfigure}
    \par\medskip
    \begin{subfigure}[t]{\linewidth}
        \centering
        \includegraphics[width=\linewidth]{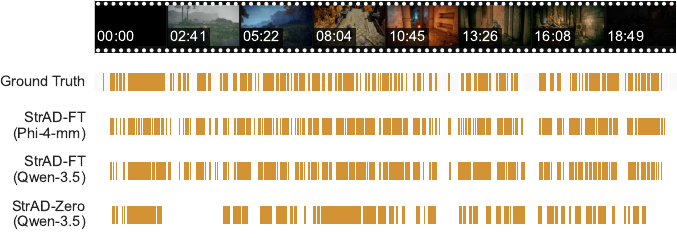}
        \caption{\emph{Elden Ring---Gameplay Preview}. Original video: \href{https://youtu.be/JldMvQMO_5U}{\nolinkurl{youtu.be/JldMvQMO_5U}}; audio-described version: \href{https://youtu.be/neQPEIAqpQ8}{\nolinkurl{youtu.be/neQPEIAqpQ8}}.}
        \label{fig:supp-timeline-b}
    \end{subfigure}
    \caption{\textbf{Full streaming output overview.} AD events for the ground truth, our fine-tuned Qwen-3.5 and Phi-4-mm \audiodescmodel{} variants, and the zero-shot \audiodesczeromodel{} baseline, placed on the full video timeline. Each bar marks the start and duration of an emitted AD.}
    \label{fig:supp-timeline}
\end{figure}

\begin{figure}[!t]
    \centering
    \includegraphics[width=\linewidth]{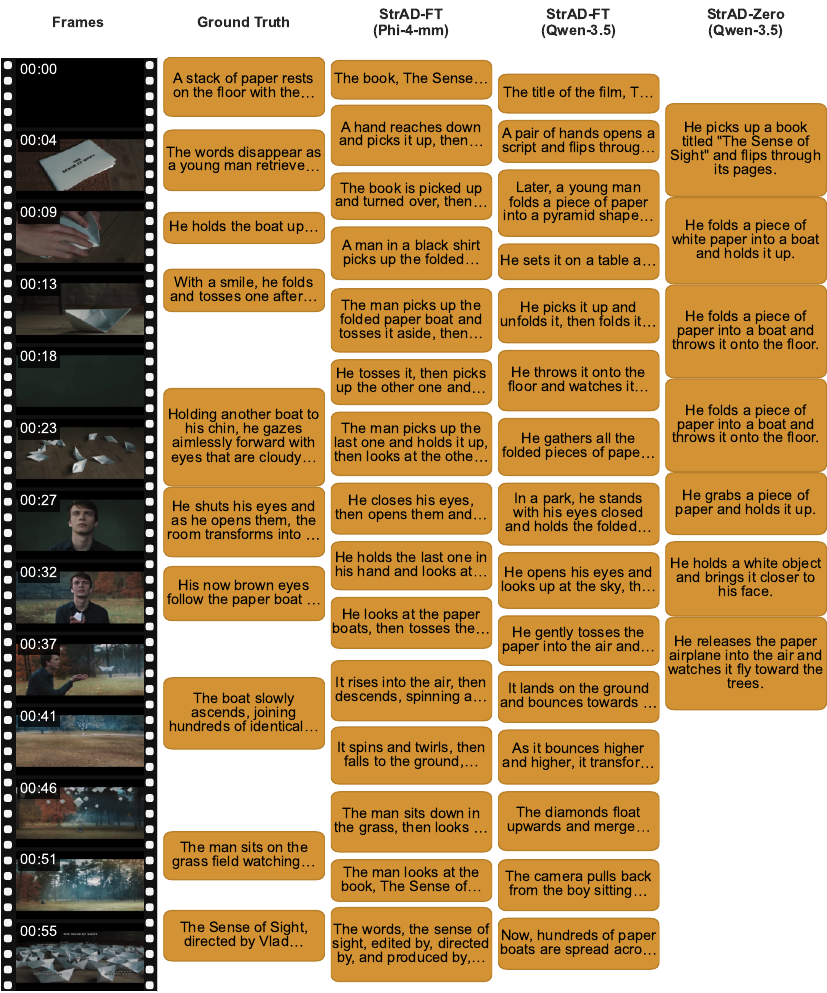}
    \caption{\textbf{Side-by-side qualitative comparison on \emph{The Sense of Sight}.} Sampled frames, the transcribed dialogue, and the AD stream emitted by the ground truth, both \audiodescmodel{} variants, and the \audiodesczeromodel{} baseline. Original video: \href{https://youtu.be/Oph8kL_Z4c4}{\nolinkurl{youtu.be/Oph8kL_Z4c4}}; audio-described version: \href{https://youtu.be/fNV8TUqjfbQ}{\nolinkurl{youtu.be/fNV8TUqjfbQ}}.}
    \label{fig:supp-comparison-a}
\end{figure}

\begin{figure}[!t]
    \centering
    \includegraphics[width=\linewidth]{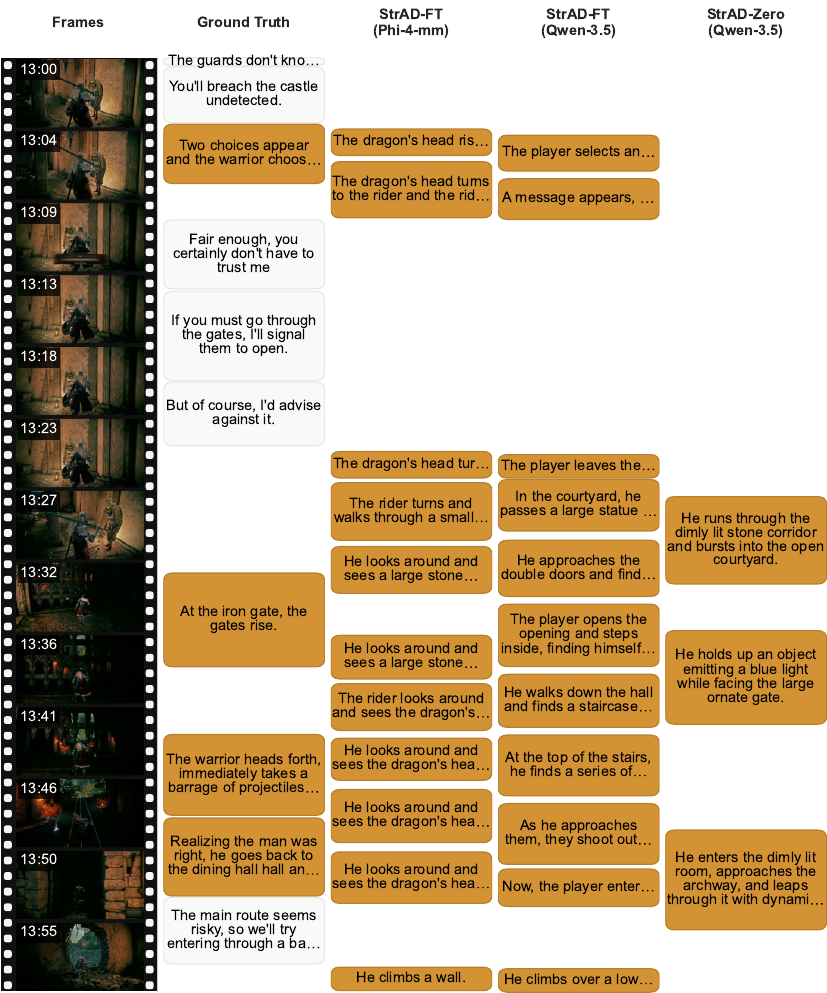}
    \caption{\textbf{Side-by-side qualitative comparison on \emph{Elden Ring---Gameplay Preview} (780--840\,s).} Sampled frames, the transcribed dialogue, and the AD stream emitted by the ground truth, both \audiodescmodel{} variants, and the \audiodesczeromodel{} baseline. Original video: \href{https://youtu.be/JldMvQMO_5U}{\nolinkurl{youtu.be/JldMvQMO_5U}}; audio-described version: \href{https://youtu.be/neQPEIAqpQ8}{\nolinkurl{youtu.be/neQPEIAqpQ8}}.}
    \label{fig:supp-comparison-b}
\end{figure}

We provide qualitative examples on two videos from \audiodescbench{}.
\cref{fig:supp-timeline} gives a full view: for each video, we plot AD events as time-aligned bars for the ground truth, both fine-tuned \audiodescmodel{} variants (Qwen-3.5 and Phi-4-mm), and the zero-shot \audiodesczeromodel{} baseline.
We observe that fine-tuned models tend to be more verbose, while zero-shot approaches produce longer AD sentences.
\cref{fig:supp-comparison-a,fig:supp-comparison-b} zoom into a 60-second segment of each video and show the corresponding video frames together with the transcribed dialogue and the AD stream produced by each model.

\section{Asset Licenses}\label{sec:appendix-licenses}

\cref{tab:licenses} lists the license of each existing asset used in this work.

\begin{table}[!t]
 \centering
 \caption{Licenses of existing assets used in this work.}
 \label{tab:licenses}
 \footnotesize
 \begin{tabular}{lll}
  \toprule
  Asset & Type & License \\
  \midrule
  Phi-4-multimodal-instruct~\citep{abouelenin2025phi}  & Model   & MIT \\ 
  Qwen-3.5~\citep{qwen3.5}                              & Model   & Apache 2.0 \\ 
  Llama-2-7B~\citep{touvron2023llama}                  & Model   & Llama 2 Community License \\ 
  Llama-3-8B~\citep{grattafiori2024llama}              & Model   & Llama 3 Community License \\ 
  VideoLLaMA2~\citep{cheng2024videollama}              & Model   & Apache 2.0 \\ 
  Whisper~\citep{radford2023robust}                    & Model   & Apache 2.0 \\ 
  Silero VAD~\citep{silero}                            & Model   & MIT \\ 
  pyannote.audio~\citep{plaquet2023pyannote}           & Model   & MIT \\ 
  CMD-AD~\citep{bain2020CMD,han2024autoad3}            & Dataset & CC BY 4.0 \\ 
  MAD~\citep{soldan2022MAD,han2023autoad}              & Dataset & Research-only (NDA) \\ 
  describealign                                        & Code    & GPL-3.0 \\ 
  vLLM~\citep{kwon2023efficient}                       & Code    & Apache 2.0 \\ 
  Optuna~\citep{akiba2019optuna}                       & Code    & MIT \\ 
  Transformers~\citep{wolf2020transformers}            & Code    & Apache 2.0 \\ 
  Accelerate~\citep{accelerate}                        & Code    & Apache 2.0 \\  
  PyTorch~\citep{paszke2019pytorch}                    & Code    & BSD-3-Clause \\ 
  WebDataset                                           & Code    & BSD-3-Clause \\ 
  \bottomrule
 \end{tabular}
\end{table}

\end{document}